\documentclass[10pt,twocolumn,letterpaper]{article}

\usepackage{cvpr}              
\PassOptionsToPackage{table,xcdraw}{xcolor}

\usepackage{colortbl}

\definecolor{cvprblue}{rgb}{0.21,0.49,0.74}
\usepackage[pagebackref,breaklinks,colorlinks,allcolors=cvprblue]{hyperref}
\usepackage{multirow}
\usepackage[table,xcdraw]{xcolor}
\usepackage{placeins}
\usepackage{float}
\usepackage{afterpage}
\usepackage{needspace}

\newtheorem{theorem}{Theorem}

\def\paperID{17992} 
\def\confName{CVPR}
\def\confYear{2025}
\title{CLIP Under the Microscope: A Fine-Grained Analysis of Multi-Object Representation}

\author{Reza Abbasi, Ali Nazari, Aminreza Sefid, Mohammadali Banayeeanzade, \\
Mohammad Hossein Rohban, Mahdieh Soleymani Baghshah\\
Sharif University of Technology, Tehran, Iran\\
{\tt\small \{reza.abbasi, ali.nazari02, aminreza.sefid, a.banayeean, rohban, soleymani\}@sharif.edu}
}

\begin{document}
\maketitle

\begin{abstract}

Contrastive Language-Image Pre-training (CLIP) models excel in zero-shot classification, yet face challenges in complex multi-object scenarios. This study offers a comprehensive analysis of CLIP's limitations in these contexts using a specialized dataset, ComCO, designed to evaluate CLIP's encoders in diverse multi-object scenarios. Our findings reveal significant biases: the text encoder prioritizes first-mentioned objects, and the image encoder favors larger objects. Through retrieval and classification tasks, we quantify these biases across multiple CLIP variants and trace their origins to CLIP's training process, supported by analyses of the LAION dataset and training progression. Our image-text matching experiments show substantial performance drops when object size or token order changes, underscoring CLIP's instability with rephrased but semantically similar captions. Extending this to longer captions and text-to-image models like Stable Diffusion, we demonstrate how prompt order influences object prominence in generated images. For more details and access to our dataset and analysis code, visit our project repository: \href{https://clip-oscope.github.io/}{https://clip-oscope.github.io/}.

\end{abstract}    
\section{Introduction}

The convergence of vision and language in artificial intelligence has led to the development of Vision-Language Models (VLMs) that can interpret and generate multimodal content. Among these, OpenAI's Contrastive Language-Image Pre-training (CLIP) model~\cite{radford2021learningtransferablevisualmodels} has been particularly influential, demonstrating remarkable capabilities in zero-shot image classification and setting new standards for multimodal understanding~\cite{Cherti_2023, gadre2023datacompsearchgenerationmultimodal, schuhmann2021laion400mopendatasetclipfiltered, thrush2022winoground}. The success of CLIP has catalyzed a wide array of applications---from image retrieval and visual question answering to text-to-image generation---signifying a paradigm shift in how models perceive and relate visual and linguistic information.

Visual Language Models like CLIP face significant challenges in understanding and reasoning about complex scenes with multiple objects and intricate relationships. CLIP struggles to identify distinct objects and model their relationships accurately, especially when captions contain the same objects but differ in their relationships. This results in difficulty distinguishing between similar captions with different object relationships. Several benchmark datasets have been introduced to elucidate the limitations of existing models in capturing subtle relational nuances. Notably, Winoground \cite{thrush2022winoground}, VL-CheckList \cite{zhao2022vl}, ARO \cite{yuksekgonul2023and}, and CREPE \cite{ma2023crepe} have been instrumental in evaluating models' capacities to accurately match images with semantically appropriate captions.

\begin{figure*}[t]
    \centering
    \includegraphics[width=\textwidth]{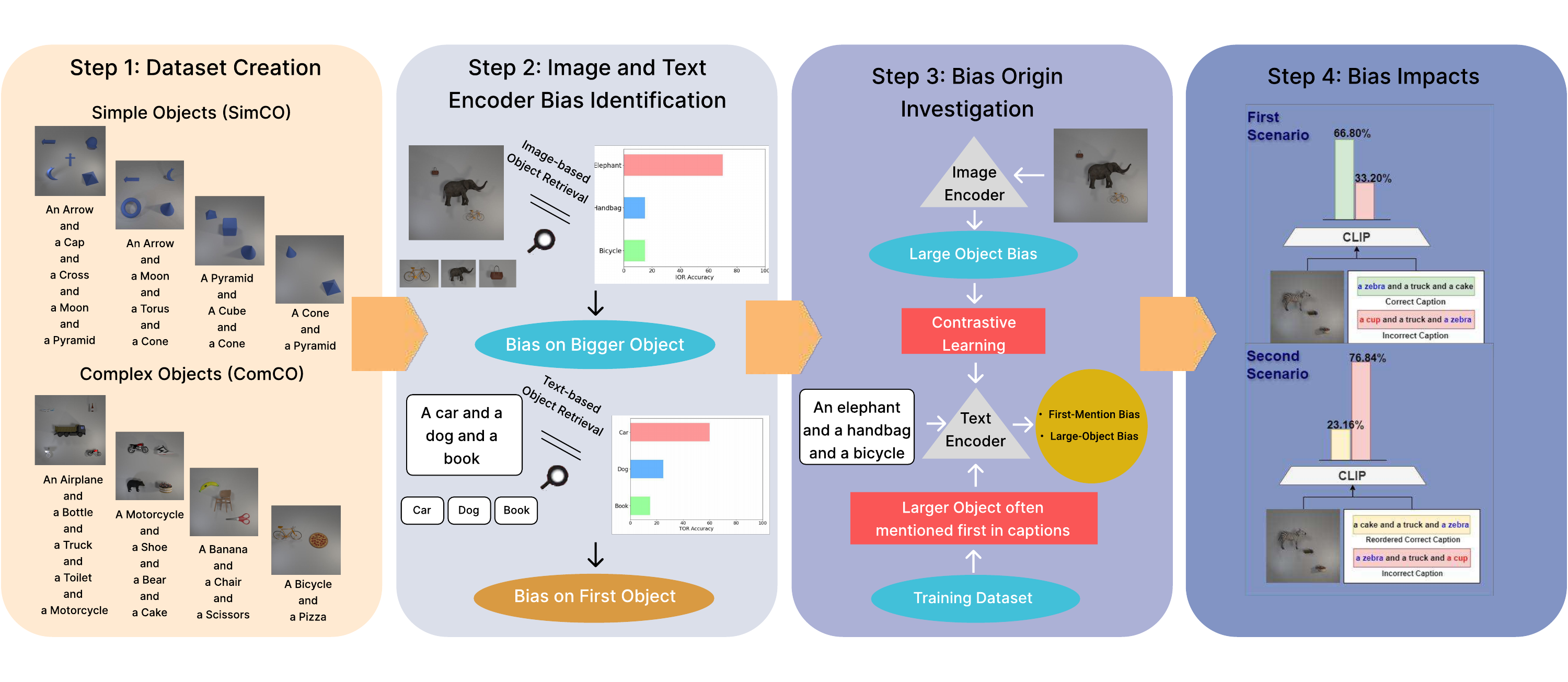}
    \caption{Overview of our key contributions. Step 1: We create ComCO dataset for controlled multi-object experiments. Step 2: We identify biases in CLIP's image encoder (favoring larger objects) and text encoder (prioritizing first-mentioned objects). Step 3: We investigate the origin of these biases, finding a connection to training data characteristics. Step 4: We demonstrate the practical impacts of these biases on image-text matching task, showing how they affect model performance in multi-object scenarios.}
    \label{fig:mainfig}
    \vspace{-0.5cm} 
\end{figure*}

Numerous studies have addressed compositionality challenges in multi-object scenarios, often through end-to-end methods like fine-tuning with hard-negative samples \cite{yuksekgonul2023and} to improve model performance. However, these approaches have faced criticism and subsequent refinement, as seen in methods like SUGARCREPE \cite{hsieh2024sugarcrepe} and \cite{sahin2024enhancing}, which generate negative captions with minor structural changes or LLMs to highlight semantic distinctions. While most focus on CLIP’s ability to distinguish structurally similar yet conceptually different captions, few studies, such as Dumpala et al. \cite{dumpala2024sugarcrepe++}, explore CLIP’s performance on semantically equivalent but structurally distinct captions, revealing a gap in understanding CLIP's inconsistency with such prompts.

While previous studies have advanced our understanding of CLIP's limitations, our work uniquely focuses on CLIP's performance with semantically equivalent but structurally varied captions rather than simply distinguishing conceptually different captions. This shift enables a deeper examination of the model’s grasp of language and visual content, where systematic errors reveal potential biases. Unlike prior works that primarily propose benchmarks or end-to-end solutions, we investigate the root causes of CLIP's behavior, delving into the mechanisms of both image and text encoders to uncover why the model displays biases and lacks robustness to certain linguistic and visual variations. To support this analysis, we introduce the \textbf{ComCO} dataset, purpose-built for examining CLIP's performance under {\it controlled} multi-object scenarios. Our study spans multiple versions of CLIP trained on diverse datasets and architectures, ensuring the broad applicability of our findings. This comprehensive approach aims to deepen our understanding of CLIP’s limitations and pave the way for more adaptable vision-language models. Beyond CLIP, our insights have significant implications for text-to-image (T2I) generative models and multimodal large language models (MLLMs), where decoding CLIP’s encoding intricacies can inform advancements in artificial intelligence across domains. As shown in Figure \ref{fig:mainfig}, our key contributions are as follows:

\begin{itemize} \item \textbf{Development of Novel Dataset}: We introduce \textit{ComCO}, a specialized dataset for creating {\it controlled} multi-object scenarios. Unlike previous benchmarks, ComCO allows control over object size and caption order, enabling precise analysis of model performance across compositional challenges and enhancing understanding of VLMs' strengths and weaknesses.

\item \textbf{Encoder Analysis}: We conduct an in-depth examination of CLIP’s image and text encoders in multi-object scenes, revealing weaknesses in preserving information for object distinction and identifying where compositional information is lost.

\item \textbf{Bias Identification}: Our study reveals that CLIP’s image encoder prefers larger objects, while the text encoder favors first-mentioned and visually larger objects, highlighting biases in CLIP's handling of visual and linguistic information.

\item \textbf{Investigation of Bias Origins}: We explore the origins of these biases, showing that larger objects are often mentioned earlier in CLIP’s training captions, and are favored in embeddings due to the abundance of their visual tokens. We substantiate this with analyses of the LAION dataset and CLIP’s training progression.

\item \textbf{Practical Impact}: We show how these biases affect performance in multi-object tasks, with significant drops in image-text matching accuracy in ComCO and COCO ~\cite{lin2015microsoftcococommonobjects}. These biases also extend to text-to-image models, influencing object prominence based on prompt order.

\end{itemize}

These findings reveal how biases in CLIP’s text and image encoders significantly reduce its performance in multi-object scenarios, emphasizing the need to address these biases to enhance vision-language models' robustness. Our work offers key insights into CLIP's behavior and lays groundwork for improving model performance in real-world applications.

\section{Methodology}
\label{sec:methodology}


\subsection{Dataset Design}
To thoroughly evaluate the performance of CLIP models in multi-object scenarios under controlled conditions, we constructed the \textbf{ComCO} (Complex COCO Objects) dataset. Utilizing Blender software allowed us precise control over the number, location, and dimensions of objects in the images (see Appendix \ref{app:dataset}).
The \textbf{ComCO} dataset comprises 72 objects derived from the COCO dataset. We generated images containing 2, 3, 4, and 5 objects. Each image is paired with a specific caption that accurately describes the objects present. This approach ensures high control over the dataset and minimizes confounding factors, providing a robust platform for evaluating the CLIP models.

We deliberately chose not to use text-to-image models for generating these datasets due to two main reasons. First, these models often lack the capability to produce high-quality, fully controlled multi-object images. Second, since CLIP is used in many of these models, utilizing them could introduce unwanted biases into our evaluations.

\subsection{Experimental Framework for Encoder Analysis}

\begin{figure*}[t]
    \centering
    \includegraphics[width=\linewidth]{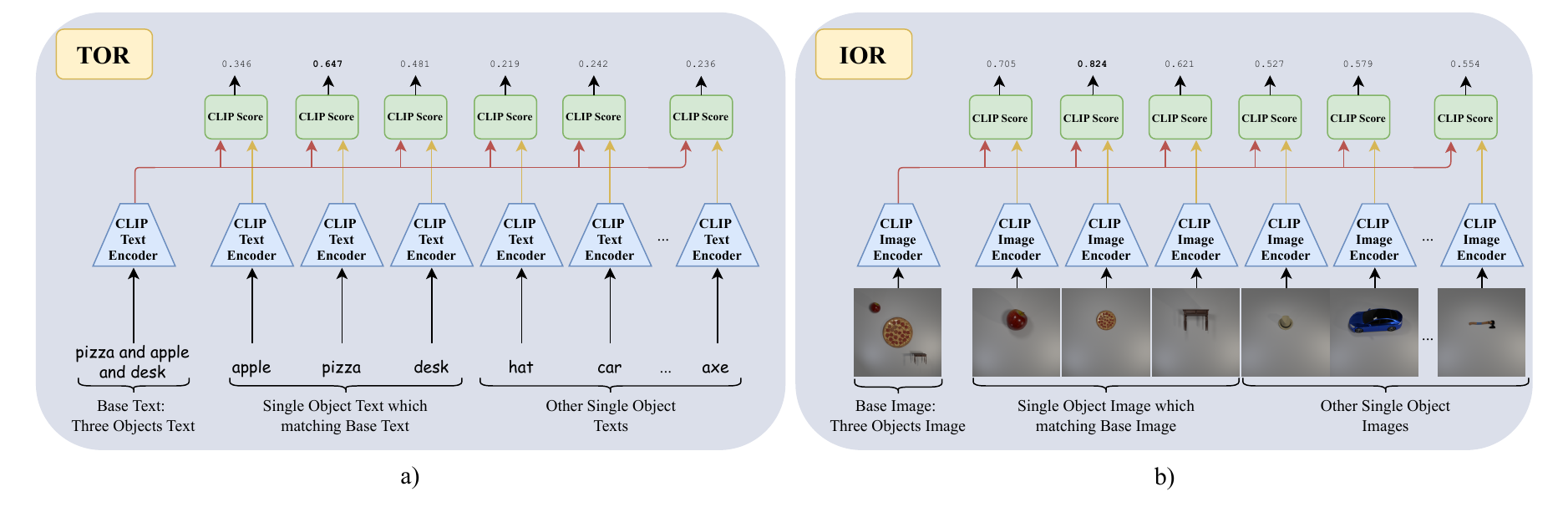}
    \caption{Experimental setup for Text-based Object Retrieval (TOR) and Image-based Object Retrieval (IOR) tasks.
    a) TOR: The CLIP text encoder generates embeddings for multi-object and single-object texts. Cosine similarity scores are calculated between the base text embedding and single-object text embeddings to identify the most similar object.
    b) IOR: The CLIP image encoder generates embeddings for multi-object and single-object images. Cosine similarity scores are calculated between the base image embedding and single-object image embeddings to identify the most similar object.
}
    \label{fig:retrievals}
        \vspace{-0.5cm} 
\end{figure*}


The main goal of this study is to evaluate the performance of CLIP's text and image encoders separately in multi-object scenarios. We aim to analyze the impact and contribution of each object in the final output of the encoders. To achieve this, we conducted experiments using our designed ComCO dataset, with images and captions containing two to five objects. To ensure the generalizability of our findings, we also validated our results on the widely-used COCO dataset. We designed two sets of experiments: retrieval-based experiments and classification-based experiments. Given the consistency of the results in both types of experiments, we have included the classification results in the appendix \ref{app:toc} and \ref{app:ioc} and explain the retrieval-based experiments bellow.

\subsubsection{TEXT-BASED OBJECT RETRIEVAL (TOR)}

The Text-based Object Retrieval task evaluates how well CLIP's text encoder can identify individual objects within multi-object captions. As illustrated in Figure \ref{fig:retrievals}a, this experiment involves several steps: First, we use CLIP's text encoder to create embeddings for both multi-object captions and single-object captions. We then measure the similarity between each multi-object caption embedding and all single-object caption embeddings. The single-object caption with the highest similarity score is considered the "retrieved" object. To assess performance, we calculate retrieval accuracy for each object position in the multi-object captions. This helps us identify any biases related to an object's position within a caption, such as favoring objects mentioned first or last.

\subsubsection{IMAGE-BASED OBJECT RETRIEVAL (IOR)}

The Image-based Object Retrieval task is similar to TOR but focuses on CLIP's image encoder. As shown in Figure \ref{fig:retrievals}b, this experiment involves several steps: We begin by using CLIP's image encoder to generate embeddings for multi-object images and single-object images. We then compute similarity scores between each multi-object image embedding and all single-object image embeddings. The single-object image with the highest similarity score is considered the "retrieved" object. To evaluate performance, we calculate retrieval accuracy for different object size categories (e.g., large, small) within the multi-object images. This allows us to determine if the image encoder shows any preference for objects of a particular size.

We also experimented with a variation of ComCO, called SimCO, where objects were replaced with simple geometric shapes from the CLEVR dataset. This was done to confirm that bias persists even with non-natural, geometric objects. Further details are provided in Appendix \ref{app:dataset}.

\section{Results and Analysis}
\label{sec:results}

Our experiments revealed significant biases in both the text and image encoders of the CLIP model. This section presents our findings, organized by encoder type and focusing on retrieval tasks. 
\subsection{Text Encoder Biases}
We observed a consistent bias in the text encoder towards the first object mentioned in descriptions. In the TOR experiment, the retrieval accuracy (as shown in Table \ref{tab:text_base_exp}) was highest for the first object, indicating its dominant influence on the overall text representation. This suggests that the text encoder prioritizes the initial object, leading to its more accurate retrieval compared to subsequent objects. The detailed results for the scenarios involving 2, 3, and 5 objects can be found in the appendix \ref{app:tor}, and experiments on longer caption templates are in Appendix \ref{app:toc-long} and \ref{app:tor-long}.

\subsection{Image Encoder Biases}

In multi-object images, the image encoder exhibited a strong bias towards larger objects. The Image-based Object Retrieval IOR experiment, detailed in Table \ref{tab:image_base_exp}, shows that larger objects were more frequently and accurately retrieved during single-object image searches. This finding highlights the image encoder's bias towards larger objects, which receive disproportionate emphasis in the final image representation.
Further detailed results, specifically for scenarios with 2, 3, and 5 objects, are provided in the appendix \ref{app:ior}.
\begin{table}[ht]
\centering
\scriptsize
\setlength{\tabcolsep}{3pt}
\renewcommand{\arraystretch}{1.2}

\caption{Performance on TOR for ComCO datasets}
\label{tab:text_base_exp}
\begin{tabular}{llcccc}
\toprule
\rowcolor[HTML]{EFEFEF}
Task & Model & \textbf{First Obj} & \textbf{Second Obj} & \textbf{Third Obj} & \textbf{Fourth Obj} \\ 
\midrule
\multirow{6}{*}{TOR}   
 & \textit{CLIP LAION} & \textbf{63.96} & 21.59 & 10.68 & 3.76 \\
 & \textit{CLIP Datacomp} & \textbf{71.13} & 16.26 & 8.74 & 3.87 \\
 & \textit{CLIP Roberta} & \textbf{44.03} & 23.73 & 18.07 & 14.18 \\
 & \textit{SIGLIP} & \textbf{58.11} & 21.16 & 10.99 & 9.73 \\
 & \textit{CLIP openAI} & \textbf{50.31} & 20.74 & 14.45 & 6.79 \\
 & \textit{NegCLIP} & \textbf{51.63} & 28.92 & 14.86 & 4.59 \\
 & \textit{SugarCrepe} & \textbf{44.29} & 30.32 & 18.73 & 6.66 \\
\bottomrule
\end{tabular}

\vspace{1em} 

\caption{Performance on IOR for ComCO datasets}
\label{tab:image_base_exp}
\begin{tabular}{llcccc}
\toprule
\rowcolor[HTML]{E4E8F2}
Task & Model & \textbf{Large Object} & \textbf{Small Obj 1} & \textbf{Small Obj 2} & \textbf{Small Obj 3} \\ 
\midrule
\multirow{6}{*}{IOR}   
 & \textit{CLIP LAION} & \textbf{85.45} & 6.36 & 5.45 & 2.73 \\
 & \textit{CLIP Datacomp} & \textbf{85.16} & 5.65 & 4.95 & 4.24 \\
 & \textit{CLIP Roberta} & \textbf{87.40} & 8.66 & 2.36 & 1.57 \\
 & \textit{SIGLIP} & \textbf{77.66} & 10.11 & 6.38 & 5.85 \\
 & \textit{CLIP openAI} & \textbf{65.22} & 17.39 & 8.70 & 8.70 \\
 & \textit{NegCLIP} & \textbf{61.67} & 15.00 & 13.33 & 10.00 \\
 & \textit{SugarCrepe} & \textbf{60.0} & 18.38 & 16.85 & 4.7 \\
\bottomrule
\end{tabular}
\end{table}

\subsection{COCO Dataset Experiments}

To validate the generalizability of our findings from the synthetic dataset, we conducted similar experiments on the COCO dataset, which comprises real images with accompanying captions. This real-world dataset allowed us to investigate whether the previously observed biases persist in more naturalistic settings.

Due to the absence of single-object images for COCO objects, we approached the IOR experiment in two ways. First, we used single-object images from the DomainNet dataset \cite{peng2019moment} as retrieval targets. Second, we introduced an alternative approach called Image-to-Text Object Retrieval (I2TOR). In I2TOR, we used the textual names of COCO objects instead of single-object images. These object names were embedded using CLIP's text encoder, allowing us to perform a retrieval task consistent with the IOR methodology while adapting to the constraints of the COCO dataset.

\begin{table}[ht]
\centering
\scriptsize
\setlength{\tabcolsep}{3pt}
\renewcommand{\arraystretch}{1.2}

\caption{Performance on TOR for coco dataset}
\label{tab:text_base_exp_coco}
\begin{tabular}{llcccc}
\toprule
\rowcolor[HTML]{EFEFEF}
Task & Model & \textbf{First Obj} & \textbf{Second Obj} & \textbf{Third Obj} & \textbf{Fourth Obj} \\ 
\midrule
\multirow{6}{*}{TOR}   
     & \textit{CLIP openAI} & \textbf{35.24} & 21.90 & 20.48 & 22.38 \\
     & \textit{CLIP LAION} & \textbf{67.89} & 13.76 & 8.26 & 10.09 \\
     & \textit{CLIP Datacomp} & \textbf{57.68} & 17.68 & 12.75 & 11.88 \\
     & \textit{CLIP Roberta} & \textbf{40.78} & 23.30 & 20.39 & 15.53 \\
     & \textit{SIGLIP} & \textbf{49.47} & 26.84 & 12.11 & 11.58 \\
     & \textit{NegCLIP} & \textbf{38.69} & 22.11 & 17.09 & 22.11 \\
\bottomrule
\end{tabular}

\vspace{1em} 

\caption{Performance on IOR for coco dataset}
\label{tab:image_base_exp_coco}
\begin{tabular}{llcccc}
\toprule
\rowcolor[HTML]{E4E8F2}
Task & Model & \textbf{Large Object} & \textbf{Small Obj 1} & \textbf{Small Obj 2} & \textbf{Small Obj 3} \\ 
\midrule
\multirow{6}{*}{IOR}   
     & \textit{CLIP openAI} & \textbf{43.02} & 28.82 & 17.13 & 11.03 \\
     & \textit{CLIP LAION} & \textbf{39.44} & 28.45 & 17.70 & 14.41 \\
     & \textit{CLIP Datacomp} & \textbf{36.71} & 29.55 & 19.13 & 14.61 \\
     & \textit{CLIP Roberta} & \textbf{36.71} & 28.61 & 19.82 & 14.86 \\
     & \textit{SIGLIP} & \textbf{36.63} & 28.29 & 20.02 & 15.06 \\
     & \textit{NegCLIP} & \textbf{44.04} & 28.86 & 16.48 & 10.62 \\
\midrule
\multirow{6}{*}{I2TOR}   
     & \textit{CLIP openAI} & \textbf{51.49} & 24.87 & 13.68 & 9.97 \\
     & \textit{CLIP LAION} & \textbf{45.50} & 27.02 & 15.91 & 11.56 \\
     & \textit{CLIP Datacomp} & \textbf{46.64} & 26.82 & 14.53 & 12.01 \\
     & \textit{CLIP Roberta} & \textbf{44.69} & 26.98 & 16.04 & 12.29 \\
     & \textit{SIGLIP} & \textbf{47.09} & 27.07 & 15.10 & 10.74 \\
     & \textit{NegCLIP} & \textbf{49.04} & 27.07 & 14.08 & 9.81 \\
\bottomrule
\end{tabular}
\end{table}


Tables \ref{tab:text_base_exp_coco} and \ref{tab:image_base_exp_coco} present the results of our COCO dataset experiments. In TOR, the first-mentioned object in COCO captions was retrieved with higher accuracy, which aligns with our earlier findings of bias in the text encoder. Similarly, in IOR, larger objects in COCO images were retrieved more accurately, consistent with the trends observed in our synthetic dataset experiments. The I2TOR results further confirmed this bias, demonstrating that even when using textual object representations, the bias towards larger objects persists.

Our experiments reveal two significant biases in the CLIP model: the text encoder shows a strong preference for the first mentioned object in textual descriptions, while the image encoder exhibits greater sensitivity to larger objects in images. These biases can significantly impact the overall system performance in various vision-language tasks, particularly in multi-object scenarios.



\section{Origin of Bias in CLIP Models}
\label{sec:origin}

\begin{figure*}[!t]
    \centering
    \includegraphics[width=0.9\linewidth]{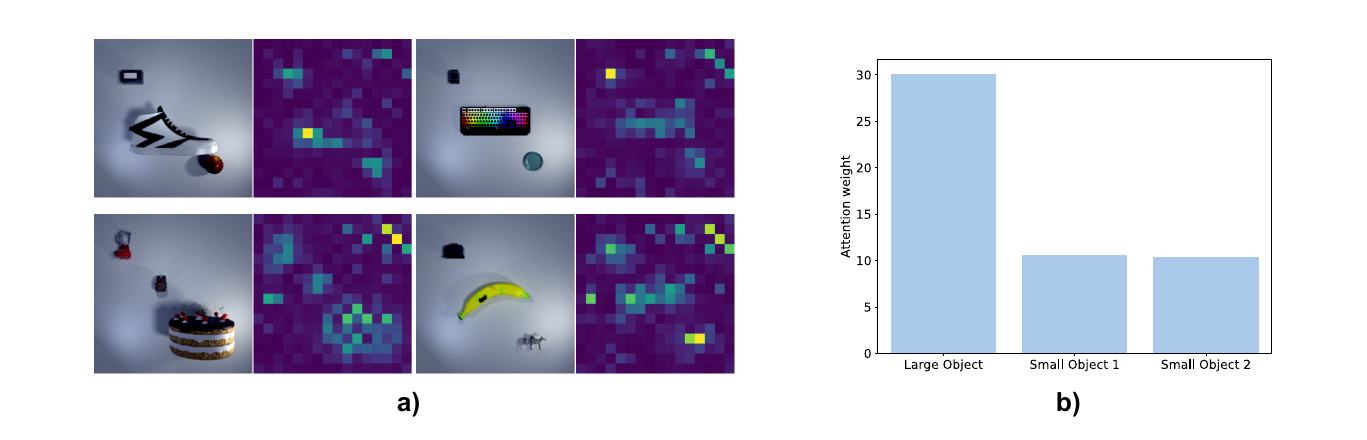}
\caption{Attention allocation from the CLS token to objects of different sizes in the ComCO dataset. a) Qualitative results showing the CLS token's attention to each object. b) Quantitative analysis of attention distribution across 8,000 images, with each image containing one large and two small objects. The bar chart shows the average attention allocated to the large object versus the smaller ones, demonstrating a bias towards larger objects.}
    \label{fig:attention_map}
\end{figure*}

In this section, we investigate the potential origins of the biases observed in CLIP models and provide evidence supporting our hypotheses.

\subsection{Bias in the Image Encoder}

The observed bias favoring larger objects within the image domain can be attributed to the architectural characteristics of Vision Transformers (ViT) \cite{alexey2020image} utilized in CLIP's image encoder. Our hypothesis is that larger objects, which occupy a greater number of patches in the ViT's patch-based image representation, exert a more significant influence on the final class (CLS) token representation. This bias is not exclusive to CLIP; it appears to be a consistent feature across ViT models, as demonstrated by our experiments detailed in the appendix.

To substantiate this hypothesis, we designed an experiment to quantify the attention allocated by the CLS token to each image patch. By calculating the cumulative attention received by each object from the CLS token, we could assess the influence of object size on attention allocation. We applied this analysis to our three-object ComCO dataset, and the results are illustrated in Figure~\ref{fig:attention_map}. The findings confirm our hypothesis: larger objects indeed receive more attention from the CLS token.

\subsection{Bias in the Text Encoder}

We explore the bias present in the text encoder from two perspectives: the attention mechanism in the model structure and the model's training method.

\subsubsection{Impact of Attention Mechanism}

Text encoder models can be categorized based on their attention mechanisms: uni-directional (causal) attention and bi-directional attention. In models with causal attention, each token attends only to preceding tokens, whereas in bi-directional models, each token attends to all tokens in the sequence.

When OpenAI introduced the CLIP model, its text encoder employed causal attention, meaning each token could only attend to tokens before it and itself. This differs from typical self-attention mechanisms, where tokens attend to all other tokens. Most CLIP models use causal self-attention, with the exception of the variant using the XLM-Roberta text encoder, which also employs self-attention. However, as shown in Table~\ref{tab:text_base_exp}, even this model exhibits the mentioned bias. This indicates that the bias does not originate from the attention mechanism itself.

\subsubsection{Role of Training Method}

\begin{table}
\centering
\scriptsize
\setlength{\tabcolsep}{3pt}
\renewcommand{\arraystretch}{1.2}
\caption{Performance on TOC and TOR for ComCO datasets}

\label{tab:diff_training_models}
\begin{tabular}{llcccc}
\toprule
\rowcolor[HTML]{EFEFEF}
Task & Model & \textbf{First Obj} & \textbf{Second Obj} & \textbf{Third Obj} & \textbf{Fourth Obj} \\ 
\midrule
\multirow{3}{*}{TOR}   
 & \textit{CLIP} & \textbf{56.28} & 22.71 & 13.17 & 7.48 \\
 & \textit{SBERT} & 29.02 & 19.80 & 17.50 & \textbf{33.57} \\
 & \textit{SimCSE} \cite{gao2021simcse} & 27.59 & 19.07 & 17.76 & \textbf{34.83} \\
\bottomrule
\end{tabular}
\vspace{-0.5cm} 
\end{table}
To determine whether the observed bias is specific to CLIP models, we compared CLIP's text encoder with two other models designed to embed sentences into a meaningful semantic space: Sentence-BERT (SBERT) \cite{reimers2019sentence} and SimCSE \cite{gao2021simcse}. The primary distinction is that CLIP's embedding space is shared between images and text, whereas SBERT and SimCSE operate solely in the text domain.

We conducted the TOR experiment on our dataset using these models. As presented in Table \ref{tab:diff_training_models}, the bias observed in CLIP differs from that in the other models. This suggests that CLIP's unique training method, which aligns images and text in a shared embedding space through contrastive learning, contributes to the bias. Therefore, to uncover the root cause of the bias, we focus on the specifics of CLIP's training procedure.

\subsection{Hypothesized Origin of Text-Side Bias in CLIP}

\begin{figure*}[t]
    \centering
    \includegraphics[width=\linewidth]{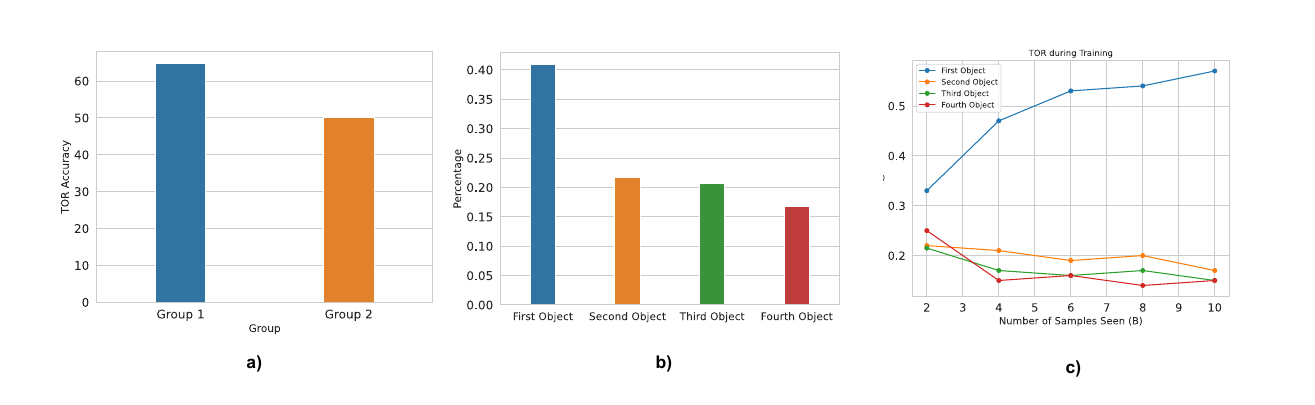}
    
\caption{a) Top-1 Object Retrieval accuracy comparison for sentences where the first object is either large or small. The higher TOR accuracy for sentences beginning with large objects supports the hypothesis that larger objects, when mentioned first, exert a stronger influence on text embeddings due to cross-modal alignment with their prominent visual representation in images. b) Distribution of the position of the largest object within image captions from the LAION datasets. The results show a consistent bias where larger objects tend to be mentioned earlier in text descriptions. c) Progression of TOR rates across different training stages, indicating that text-side bias strengthens as the model is exposed to more data, suggesting the cumulative effect of image-side bias being transferred to the text encoder through contrastive learning.}
    \label{fig:claims}
\end{figure*}

We hypothesize that the text-side bias in CLIP, which favors objects mentioned earlier in text descriptions, originates from the image-side bias toward larger objects and is transferred to the text encoder during contrastive training. We present evidence supporting this hypothesis through two key claims and an analysis of the training progression.

\paragraph{Claim 1: Larger Objects Have More Influence on Text Embeddings.}
Building upon the established image-side bias discussed earlier, we posit that objects with larger physical sizes exert more influence on CLIP's text embeddings due to the alignment enforced during contrastive training. To test this, we categorized objects in the DomainNet dataset into large, medium, and small groups based on their relative physical sizes in real-world (with the full list of objects provided in the appendix \ref{app:categorized_domainnet}). Specifically, objects smaller than a school bag were categorized as small, objects sized between a school bag and a medium-sized car were classified as medium, and objects larger than a car—up to significantly larger items—were considered large. We then constructed two sets of sentences, each containing four objects: one set with a large object mentioned first followed by three medium-sized objects, and another with a small object mentioned first followed by three medium-sized objects. 

Figure \ref{fig:claims}.a compares the TOR accuracy for the first object in these two groups. The higher TOR accuracy for sentences beginning with large objects supports our hypothesis that larger objects, when mentioned first, have a more significant impact on the text embeddings due to the cross-modal alignment with their prominent representation in images.

\paragraph{Claim 2: Caption Bias in Training Datasets.} To investigate potential biases in CLIP's training data, we analyzed both the LAION \cite{schuhmann2022laion} and COCO datasets. Due to limited computational resources and the large size of the LAION dataset, which contains over 2 billion image-text pairs, we randomly selected a subset of 200,000 samples for our analysis. Using the Llama3 model, we extracted objects from the image captions and employed the Language Segment-Anything tool to generate object masks in the corresponding images, calculating their areas based on these masks. A detailed description of our LAION dataset analysis methodology can be found in Appendix \ref{subsec:laion_analysis}.

Figure\ref{fig:claims}.b shows the position of the largest object within each caption. The results indicate that, in the majority of cases, the largest object in an image is mentioned earlier in its caption. The same experiment was conducted on the COCO dataset, with detailed results and the distribution for two to five object scenarios provided in Appendix \ref{app:coco-anlysis}. This demonstrates a consistent bias in the training data, where larger objects are not only more visually prominent but are also described earlier in text annotations.

\paragraph{Analysis of Bias Development During Training.}
To further validate our hypothesis, we examined the progression of text-side bias during CLIP's training. We utilized model checkpoints from the LAION dataset at five training stages, corresponding to exposure to 2, 4, 6, 8, and 10 billion samples. We conducted TOR experiments at each stage, focusing on the retrieval accuracy for the first object mentioned in text descriptions.

Figure\ref{fig:claims}.c depicts the evolution of the TOR rate across different training stages for scenarios with varying numbers of objects (from 3 to 8). The consistent upward trend in the TOR rate as the model is exposed to more training data suggests that the text-side bias strengthens over time, likely due to the cumulative effect of the image-side bias being transferred to the text encoder through contrastive learning.

\paragraph{Incomplete Text Representation of CLIP}

Here we want to theoretically highlight why the CLIP text encoder could learn an incomplete representation of the text. 
Let $\mathbf{z}$ and $\mathbf{w}$ represent a latent representation of an image content described in the caption, and such visual content not mentioned in the text, respectively. For example, $\mathbf{z}$ represents the fact that an image contains ``a horse that is eating the grass.'' In this case, $\mathbf{w}$ might represent other details in the image, like the ``horse color,'' ``where the horse is located,'' etc. We assume a data generative process as follows:
\begin{align*}
    & I := g({\mathbf z}, {\mathbf w}) \\
    & T := h({\mathbf z}),
\end{align*}
where $I$ is the image, and $T$ is its corresponding caption. 

Now we want to learn a joint embedding of the image and text through the CLIP. Here, we assume that $f_\theta(.)$ and $i_\omega(.)$ as learnable functions that map the image and text into the joint embedding space, respectively. 
\begin{theorem}
Let elements of ${\mathbf z}$ be independent, zero-mean, and unit-variance. The contrastive loss for the ideal text encoder, $i_\omega(T) = {\mathbf z}$ converges to that of a non-ideal incomplete one, i.e. $i_{\omega^\prime}(T) = {\mathbf z}_s$, where ${\mathbf z}_s$ is the first $d-k$ dimensions of   ${\mathbf z}$, with $k$ being a constant, and $d \rightarrow \infty$. 
\end{theorem}

Proof: The contrastive loss in making this learning happen can be written as:
\begin{align}
    \mathbb{E}_{\mathbf{z}, \mathbf{z}^\prime, \mathbf{w}} \Bigg\{ 
    \frac{\exp(sim(\mathbf{z}, \mathbf{z}))}{\exp(sim(\mathbf{z}, \mathbf{z})) + \sum_k \exp(sim(\mathbf{z}, \mathbf{z}^\prime_k))} 
    \Bigg\}
\end{align}

with 
$$
sim(\mathbf{z}, \mathbf{z}') = S(f_\theta(g(\mathbf{z}, \mathbf{w}), i_\omega(h(\mathbf{z}')))),
$$
and $\mathbf{z}$ and $\{\mathbf{z}^\prime_k | 1 \leq k \leq b\} $ are $b+1$  i.i.d. samples of the content in the representation space, and $S$ is some normalized similarity metric, e.g. cosine similarity, and $b+1$ is the batch size. We assume that elements of ${\mathbf z}$ are independent, unit-variance, and zero mean. We further assume that the dimensionality of $\mathbf z$, denoted as $d$, goes to infinity. 

Under such conditions, and based on Law of Large Numbers, $\| {\mathbf z} \| \xrightarrow{p} \sqrt{d}$, when $d$ is large. Therefore, for any two independent copies of $\mathbf z$, ${\mathbf z}^\prime_k$, we have $sim({\mathbf z}, {\mathbf z}^\prime_k) = {\mathbf z}^\top {\mathbf z}^\prime_k / (\| {\mathbf z} \| \| {\mathbf z}^\prime_k \|) \xrightarrow{p} 0.$

It is evident that in the ideal case, $f_\theta(g(\mathbf{z}, \mathbf{w})) = \mathbf{z}$ and also $i_\omega(h(\mathbf{z})) = \mathbf{z}$, so the contrastive loss would converge to $e/(e + b)$, as the numerator is $e$, and the second term in the denominator converges to $\exp(0) = 1$, according to the Mann-Wald's theorem. 

\begin{figure*}[t]
    \centering
    \includegraphics[width=0.9\linewidth]{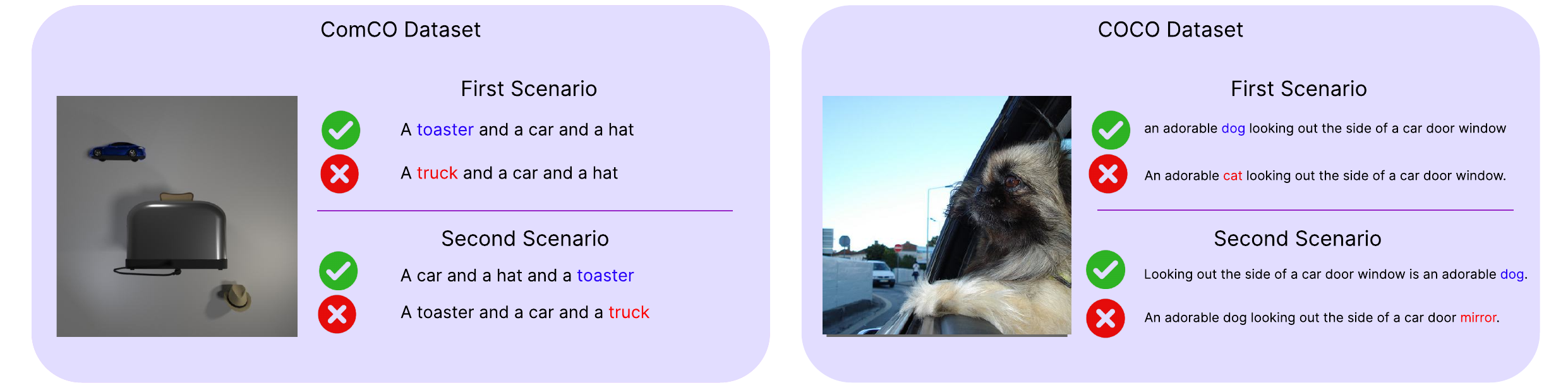}
    \caption{An example of the correct and incorrect caption structures in the first and second scenarios.}
    \label{fig:image-text-match}
    \vspace{-0.2cm}
\end{figure*}
However, we show that other learning of this representation could achieve the same amount of loss. For instance, let $\mathbf{z}_s$ be the first $d-k$ elements of $\mathbf{z}$, with $k$ being a {\it constant}. We show that if $f_{\theta^\prime}(I) = \mathbf{z}_s$ and $i_{\omega^\prime}(T) = \mathbf{z}_s$, the same loss would be achieved in the limit of large $d$. To see this, note that the numerator stays the same, i.e. $e$, while the second term in the denominator still converges to $b \exp(0) = b$.

This means that even if the image and text encoder of the CLIP only partially recover the content embedding, they reach an excellent loss. But such possible incomplete representations of $\mathbf{z}$ are combinatorially large, making convergence of the CLIP to such local minima pretty likely. This makes the text encoding of CLIP be far from ideal. Furthermore, the text encoder would become {\it biased}, depending on which of such local minima it converges to. Based on this explanation, we would expect a text encoder that has learned a complete representation to exhibit such biases to a lesser degree. As mentioned earlier, the subject of learning text representations in VLMs that are discriminative of hard negatives (e.g. NegCLIP) has been around for few years. We tested one of strongest such models, \cite{hsieh2024sugarcrepe}, in our benchmark to validate the hypothesis that an incomplete text representation is one of the causes of the bias in the VLMs. We noticed that this model shows lower bias based on our benchmark (see the SugarCrepe model in  tables \ref{tab:text_base_exp} and \ref{tab:image_base_exp}).

We have developed an initial approach to address the identified bias in the CLIP model, which is presented in Appendix \ref{sec:pre-method}. While this method is specific to our current dataset, it represents a promising step toward addressing these challenges and can inspire further advancements. This work demonstrates our commitment to exploring practical solutions while maintaining the primary focus of this study on the analysis of bias and its implications.

\section{Practical Impacts of Encoder Biases}
The biases observed in CLIP's image and text encoders significantly impact model performance in real-world applications. This section explores how these biases manifest in image-text matching tasks, while further analyses of text-to-image generation impacts are presented in Appendix \ref{sec:appendix_text_to_image}.

Our analysis in this section serves two primary purposes. First, it provides concrete evidence of how these theoretical biases can translate into practical limitations. Second, it offers insights into potential areas for improvement in vision-language models, particularly in handling complex, multi-object scenarios. Through a series of carefully designed experiments, we illustrate how the biases in both text and image encoders can lead to unexpected or suboptimal results in tasks that are crucial for many downstream applications.
\subsection{Image-Text Matching}
\label{sec:impact}

Building upon our findings of biases in CLIP’s image and text encoders, we now demonstrate how these biases tangibly affect the model’s performance in image-caption matching tasks. We designed two experimental scenarios, conducted on both the ComCO and COCO datasets, to evaluate these biases. The results of these experiments are summarized in Table \ref{table:performance_drop}. To better illustrate the differences between these two scenarios, an example of the caption structures is shown in Figure \ref{fig:image-text-match}.
In each scenario, we created incorrect captions by switching one object in the caption with an object that is not present in the image. Additionally, GPT-4O \cite{achiam2023gpt} was used to rewrite the captions in the COCO dataset.

\paragraph{First Scenario}
In the first scenario, biases assist the model in distinguishing between the correct and incorrect captions. In the correct captions, the largest object in the image is placed at the beginning, aligning with the model’s bias towards prioritizing first-mentioned objects and larger objects. For the incorrect captions, the non-existent object is deliberately placed at the beginning, which helps the model recognize the difference between the correct and incorrect captions more effectively. This positioning emphasizes the discrepancy early on, allowing the model to better detect the mismatch between the caption and the image. The performance of different models in this scenario can be seen in Table \ref{table:performance_drop} under the "First Scenario" column.

\paragraph{Second Scenario}
In the second scenario, biases lead the model to make errors. The correct captions place the largest object at the end of the sentence, disrupting the model’s bias towards objects mentioned earlier and its preference for larger objects. In the incorrect captions, the non-existent object is placed at the end, making it more difficult for the model to differentiate between correct and incorrect captions as its attention is drawn away from the critical discrepancies. The performance of different models in this scenario is shown in Table \ref{table:performance_drop} under the "Second Scenario" column.

\begin{table}[ht]
\centering
\scriptsize
\setlength{\tabcolsep}{5pt} 
\renewcommand{\arraystretch}{1.2} 

\caption{Performance Comparison on Image-Text Matching for ComCO and COCO Datasets}
\label{table:performance_drop}
\begin{tabular}{l l c c}
\toprule
\rowcolor[HTML]{EFEFEF}
\textbf{Dataset} & \textbf{Model} & \textbf{First Scenario} & \textbf{Second Scenario} \\ 
\midrule

\multirow{6}{*}{ComCO} 
    & \textit{CLIP Datacomp} \cite{gadre2024datacomp} & \textbf{99.99} & 67.50 \\
    & \textit{CLIP Roberta} & \textbf{99.98} & 64.75 \\
    & \textit{SIGLIP} \cite{zhai2023sigmoid} & \textbf{99.49} & 72.36 \\
    & \textit{CLIP openAI} & \textbf{99.59} & 52.23 \\
    & \textit{NegCLIP} & \textbf{96.82} & 46.94 \\
    & \textit{SugarCrepe} & \textbf{98.55} & 60.43 \\
\midrule

\multirow{6}{*}{COCO} 
    & \textit{CLIP Datacomp} \cite{gadre2024datacomp} & \textbf{71.2} & 54.2 \\
    & \textit{CLIP Roberta} & \textbf{72.2} & 54.1 \\
    & \textit{SIGLIP} \cite{zhai2023sigmoid} & 64.8 & 39.5 \\
    & \textit{CLIP openAI} & \textbf{63.5} & 26.4 \\
    & \textit{NegCLIP} & \textbf{72} & 28.7 \\
    & \textit{SugarCrepe} & \textbf{80.0} & 40.9 \\

\bottomrule
\end{tabular}
\end{table}

By comparing these two scenarios, we demonstrate that biases in CLIP can either help or hinder the model’s performance depending on how captions are structured. The experimental results, particularly with the use of GPT-4O for caption rephrasing in the COCO dataset, reveal how such biases can influence the accuracy of image-text matching tasks. These biases must be addressed to improve CLIP’s robustness in real-world multi-object scenarios.

For further insights on how these biases affect text-to-image generation, refer to our extended experiments in Appendix \ref{sec:appendix_text_to_image}.

\section{Conclusion}

This study uncovers biases in CLIP’s encoders, with the text encoder favoring first-mentioned objects and the image encoder emphasizing larger ones, which impacts performance in multi-object tasks. Using the ComCO dataset, we highlighted these biases' effects on object representation and positioning, underscoring the need for balanced training. We attribute these biases to CLIP's contrastive framework, where alignment issues propagate across modalities. Addressing these biases is essential for vision-language advancements, as seen with models like Stable Diffusion. Future work should explore counterfactual data augmentation and attention regularization to reduce such biases.



{
    \small
    \bibliographystyle{ieeenat_fullname}
    \bibliography{main}
}

\clearpage

\setcounter{page}{1}

\onecolumn
{
    \centering
    \Large
    \textbf{\thetitle}\\
    \vspace{0.5em}Supplementary Material \\
    \vspace{1.0em}
}

\section{Appendix}

\subsection{The SIMCO and ComCO Datasets}
\label{app:dataset}

\subsubsection{The SIMCO Dataset}
The SIMCO dataset comprises 17 objects. These 17 objects are:

\begin{center}
\large 
\begin{tabular}{@{} l @{\hspace{4em}} l @{\hspace{4em}} l @{}}
Cube & Sphere & Cylinder \\
Mug & Pentagon & Heart \\
Cone & Pyramid & Diamond \\
Moon & Cross & Snowflake \\
Leaf & Arrow & Star \\
Torus & Pot &
\end{tabular}
\end{center}

\begin{figure*}[htbp]
    \centering
    \includegraphics[width=0.9\linewidth]{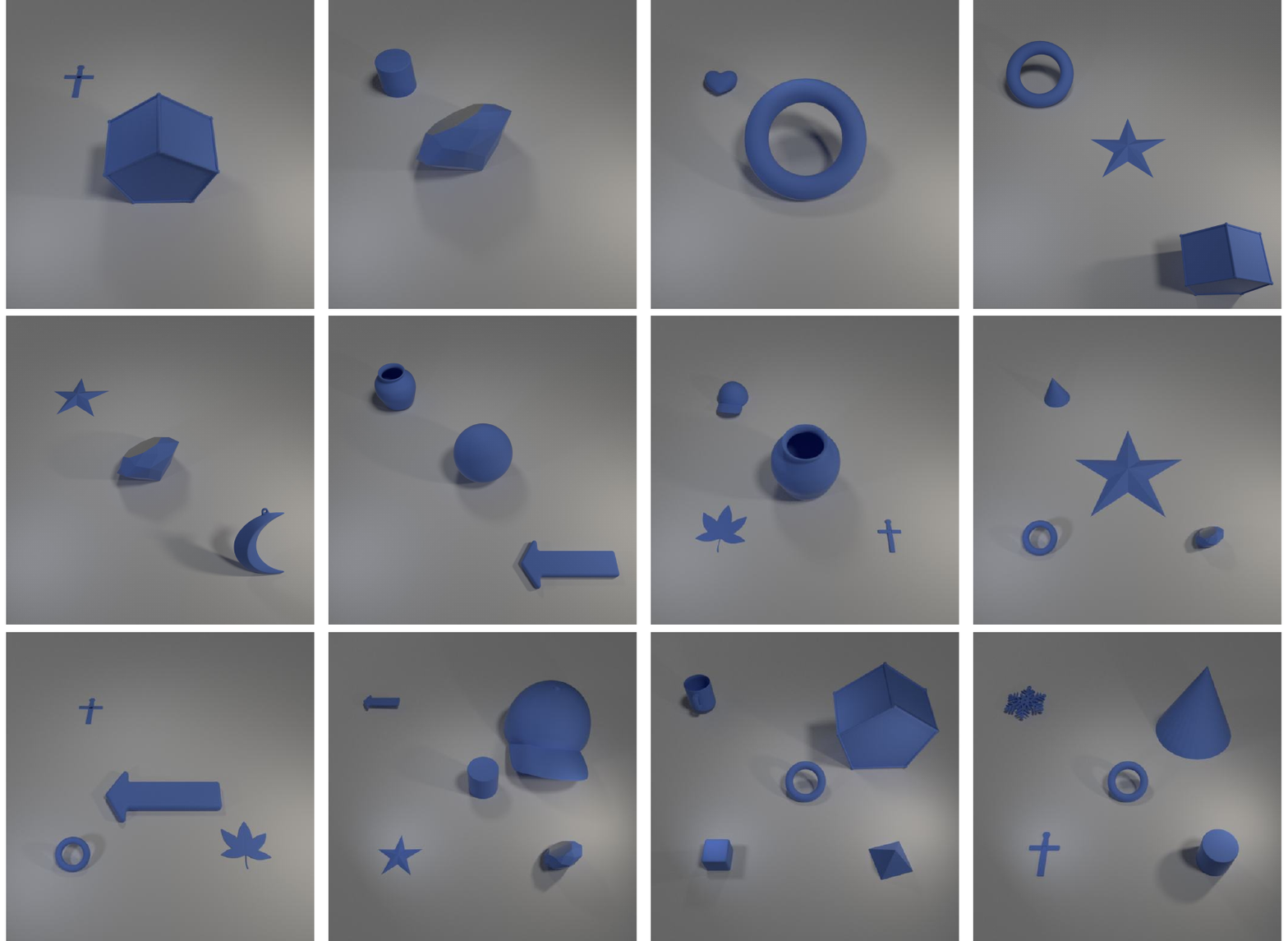}
    \caption{Examples of the SimCO dataset}
    \label{fig:simco}
\end{figure*}

Using Blender software, a collection of images containing 2 to 5 objects has been created from these 17 objects. The total number of images in this dataset is approximately 85,000. Examples of these images can be seen in Figure \ref{fig:simco}.

\FloatBarrier  
\clearpage  

\subsubsection{The ComCO Dataset}

The ComCO dataset contains 72 objects, as listed below:
\begin{center}
\large 
\begin{tabular}{@{} l @{\hspace{2em}} l @{\hspace{2em}} l @{\hspace{2em}} l @{\hspace{2em}} l @{\hspace{2em}} l @{}}
person & bicycle & car & motorcycle & airplane & bus \\
train & truck & boat & traffic light & fire hydrant & street sign \\
stop sign & parking meter & bench & bird & cat & dog \\
horse & sheep & cow & dining table & cell phone & elephant \\
bear & zebra & giraffe & hat & backpack & umbrella \\
shoe & eye glasses & handbag & tie & suitcase & frisbee \\
skis & snowboard & kite & baseball bat & baseball glove & tennis racket \\
wine glass & hot dog & potted plant & teddy bear & hair drier & hair brush \\
skateboard & surfboard & bottle & plate & cup & fork \\
knife & spoon & bowl & banana & apple & sandwich \\
orange & broccoli & carrot & pizza & donut & cake \\
chair & couch & bed & mirror & window & desk \\
toilet & door & tv & laptop & mouse & remote \\
keyboard & microwave & oven & toaster & sink & refrigerator \\
blender & book & clock & vase & scissors & toothbrush \\
\end{tabular}
\end{center}

In this dataset, a collection of images containing 2 to 5 different objects has also been generated. The total number of images in this dataset is approximately 190,000. Various examples from this dataset can be seen in Figure \ref{fig:comco}.

\begin{figure*}[htbp]
    \centering
    \includegraphics[width=0.9\linewidth]{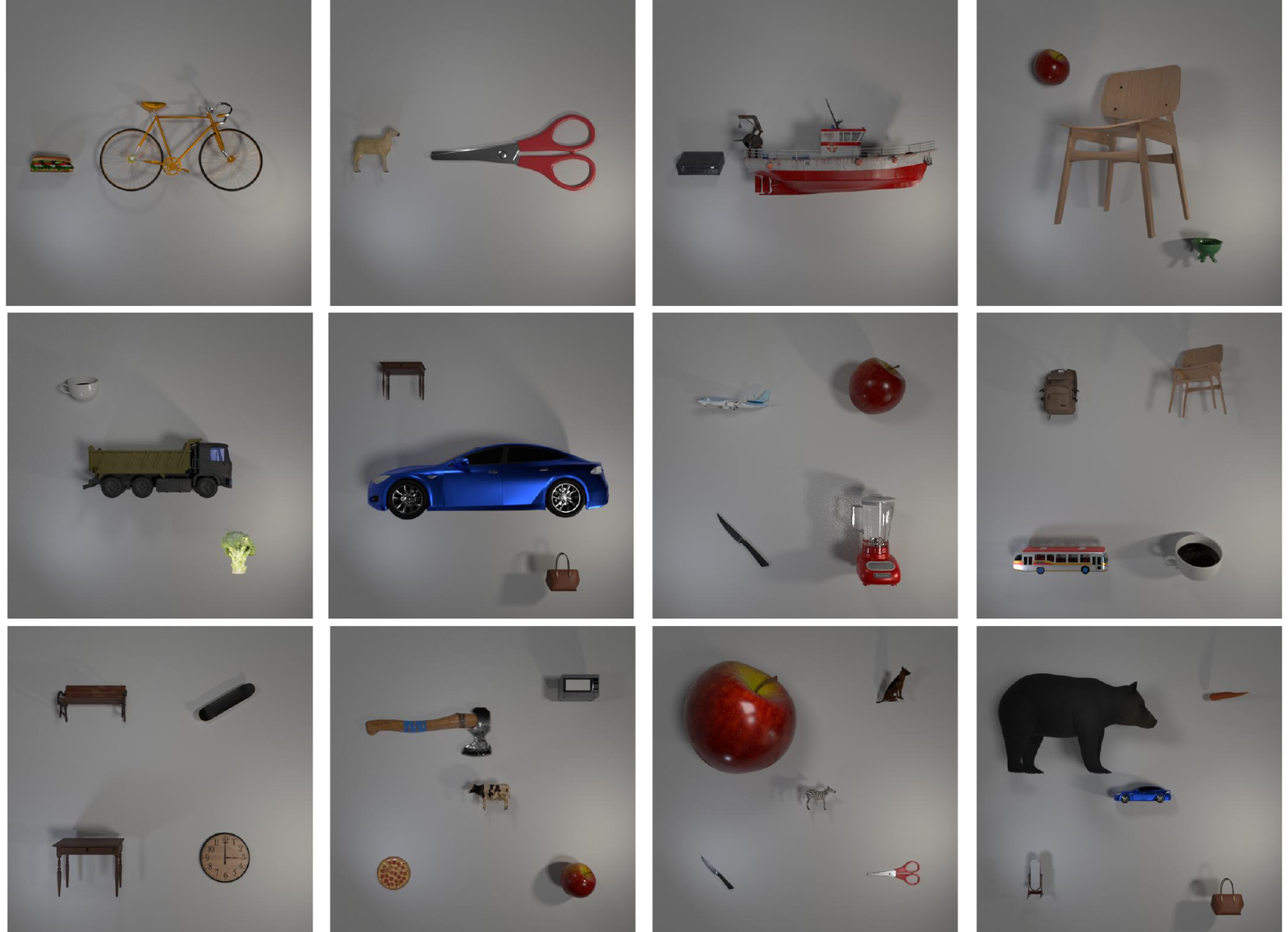}
    \caption{Examples of the ComCO dataset}
    \label{fig:comco}
\end{figure*}

\FloatBarrier  

\subsection{Text-based Object Classification}
\label{app:toc}
\subsubsection{Objective}
The Text-based Object Classification experiment was designed to evaluate CLIP's text encoder's ability to represent individual objects within multi-object captions. Our goal was to quantify any potential bias in the representation of objects based on their position in the text.
\begin{figure}[H]
    \centering
    \includegraphics[width=0.8\linewidth]{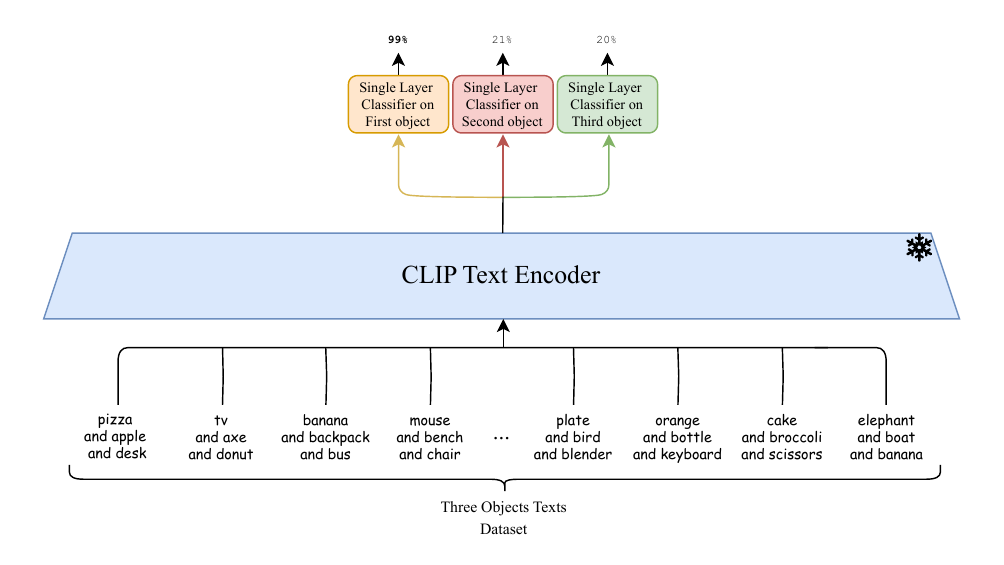}
    \caption{Illustration of the Text-based Object Classification experiment. The figure demonstrates how embeddings are calculated for multi-object captions using CLIP's text encoder. A single-layer classifier is then trained on these embeddings to classify individual objects.}
    \label{fig:toc_main_fig}
\end{figure}

\subsubsection{Methodology}

\begin{enumerate}
    \item \textbf{Dataset Preparation}: 
    \begin{itemize}
        \item We used both the SimCO and ComCO datasets, which contain captions describing scenes with 2 to 5 objects.
        \item Each caption in the dataset follows a consistent format: ``Object1 and Object2 and ... and ObjectN''.
    \end{itemize}

    \item \textbf{Text Embedding Generation}:
    \begin{itemize}
        \item For each multi-object caption, we used CLIP's text encoder to generate a text embedding.
        \item This embedding is a high-dimensional vector representation of the entire caption.
    \end{itemize}

    \item \textbf{Classifier Training}:
    \begin{itemize}
        \item For each object position (1st, 2nd, 3rd, etc.), we trained a separate single-layer classifier.
        \item Input: The text embedding of the multi-object caption.
        \item Output: The predicted object class for that specific position.
    \end{itemize}

    \item \textbf{Evaluation}:
    \begin{itemize}
        \item We tested each classifier on a held-out portion of the dataset.
        \item For each caption, we recorded whether the classifier correctly identified the object at its respective position.
        \item We calculated the classification accuracy for each object position across all test captions.
    \end{itemize}

\end{enumerate}

We conducted the TOC experiment on various models under different scenarios, and the results are presented in Table \ref{tab:toc_total}. This experiment was repeated on both the SIMCO and ComCO datasets.

\begin{table*}[htbp]
    \centering
    \scriptsize
    \setlength{\tabcolsep}{4pt}
    \renewcommand{\arraystretch}{1.1}
    \caption{Text-based Object Classification}
    \label{tab:toc_total}
    \begin{tabular}{lllccccc}
    \toprule
    \rowcolor[HTML]{E4E8F2}
    Number of Objects & Dataset & Model & \textbf{First Object} & \textbf{Second Object} & \textbf{Third Object} & \textbf{Fourth Object} & \textbf{Fifth Object} \\ 
    \midrule
    \multirow{16}{*}{n = 2}  & \multirow{8}{*}{SimCO}    
    &  \textit{ViT-H-14 (DFN)}& 99.86 & 97.09 & - & - & - \\
    & & \textit{ViT-SO400M-SigLIP} & 98.67 & 91.29 & - & - & - \\
    & & \textit{ViT-L-14 (datacomp)}& 99.76 & 96.77 & - & - & - \\
    & & \textit{xlm-roberta-large-ViT-H-14}& 99.03 & 89.87 & - & - & - \\
    & & \textit{ViT-L-14 (laion2b)} & 99.70 & 97.57 & - & - & - \\
    & & \textit{ViT-L-14 (openai)} & 97.62 & 91.30 & - & - & - \\
    & & \textit{ViT-B-32 (openai)} & 96.85 & 73.00 & - & - & - \\
    & & \textit{NegCLIP} & 98.19 & 84.43  & - & - & - \\

    \cmidrule{2-8}
    
    & \multirow{8}{*}{ComCO}    
    &  \textit{ViT-H-14 (DFN)}& 99.90 & 96.56 & - & - & - \\
    & & \textit{ViT-SO400M-SigLIP} & 98.47 & 93.18 & - & - & - \\
    & & \textit{ViT-L-14 (datacomp)}& 99.74 & 96.86 & - & - & - \\
    & & \textit{xlm-roberta-large-ViT-H-14}& 99.16 & 91.57 & - & - & - \\
    & & \textit{ViT-L-14 (laion2b)} & 99.72 & 96.24 & - & - & - \\
    & & \textit{ViT-L-14 (openai)} & 97.93 & 96.69 & - & - & - \\
    & & \textit{ViT-B-32 (openai)} & 96.86 & 85.42 & - & - & - \\
    & & \textit{NegCLIP} & 99.30 & 92.09 & - & - & - \\
    
    \midrule 
    \multirow{16}{*}{n = 3} & \multirow{8}{*}{SimCO}   
    &  \textit{ViT-H-14 (DFN)}& 99.46 & 60.47 & 76.99 & - & - \\
    & & \textit{ViT-SO400M-SigLIP} & 98.23 & 71.42 & 45.80 & - & - \\
    & & \textit{ViT-L-14 (datacomp)}& 99.49 & 45.80 & 78.66 & - & - \\
    & & \textit{xlm-roberta-large-ViT-H-14}& 99.26 & 49.08 & 64.07 & - & - \\
    & & \textit{ViT-L-14 (laion2b)} & 98.93 & 56.87 & 72.37 & - & - \\
    & & \textit{ViT-L-14 (openai)} & 91.87 & 50.75 & 68.38 & - & - \\
    & & \textit{ViT-B-32 (openai)} & 92.55 & 38.61 & 52.94 & - & - \\
    & & \textit{NegCLIP} & 95.80 & 44.70 & 59.11 & - & - \\

    \cmidrule{2-8}
    & \multirow{8}{*}{ComCO}    
    &  \textit{ViT-H-14 (DFN)}& 99.73 & 59.80 & 73.63 & - & - \\
    & & \textit{ViT-SO400M-SigLIP} & 96.94 & 70.26 & 29.28 & - & - \\
    & & \textit{ViT-L-14 (datacomp)}& 99.53 & 45.13 & 74.15 & - & - \\
    & & \textit{xlm-roberta-large-ViT-H-14}& 99.20 & 53.34 & 57.15 & - & - \\
    & & \textit{ViT-L-14 (laion2b)} & 99.26 & 58.58 & 64.74 & - & - \\
    & & \textit{ViT-L-14 (openai)} & 90.86 & 49.67 & 83.49 & - & - \\
    & & \textit{ViT-B-32 (openai)} & 87.97 & 45.77 & 63.13 & - & - \\
    & & \textit{NegCLIP} & 56.94 & 98.03 & 56.66 & - & - \\
     
    \midrule
    \multirow{16}{*}{n = 4} & \multirow{8}{*}{SimCO}   
    
    &  \textit{ViT-H-14 (DFN)}& 99.46 & 34.57 & 36.73 & 62.35 & - \\
    & & \textit{ViT-SO400M-SigLIP} & 98.23 & 69.91 & 26.10 & 6.54 & - \\
    & & \textit{ViT-L-14 (datacomp)}& 99.00 & 23.76 & 35.55 & 60.55 & - \\
    & & \textit{xlm-roberta-large-ViT-H-14}& 99.26 & 27.97 & 28.84 & 48.34 & - \\
    & & \textit{ViT-L-14 (laion2b)} & 98.82 & 34.21 & 31.41 & 54.73 & - \\
    & & \textit{ViT-L-14 (openai)} & 90.48 & 35.19 & 30.50 & 59.29 & - \\
    & & \textit{ViT-B-32 (openai)} & 90.76 & 22.77 & 25.36 & 40.45 & - \\
    & & \textit{NegCLIP} & 96.50 & 9.33 & 4.79 & 15.58 & - \\

    \cmidrule{2-8}
    & \multirow{8}{*}{ComCO}    
    &  \textit{ViT-H-14 (DFN)}& 99.76 & 31.74 & 35.29 & 54.82 & - \\
    & & \textit{ViT-SO400M-SigLIP} & 97.27 & 72.51 & 33.25 & 5.79 & - \\
    & & \textit{ViT-L-14 (datacomp)}& 99.46 & 22.82 & 32.93 & 58.18 & - \\
    & & \textit{xlm-roberta-large-ViT-H-14}& 99.60 & 26.27 & 26.20 & 36.51 & - \\
    & & \textit{ViT-L-14 (laion2b)} & 98.89 & 31.64 & 20.90 & 47.76 & - \\
    & & \textit{ViT-L-14 (openai)} & 87.17 & 30.60 & 31.69 & 74.49 & - \\
    & & \textit{ViT-B-32 (openai)} & 88.24 & 24.23 & 28.30 & 49.82 & - \\
    & & \textit{NegCLIP} & 98.73 & 28.05 & 30.83 & 43.82 & - \\
     
    \midrule
    \multirow{16}{*}{n = 5}& \multirow{8}{*}{SimCO}

    &  \textit{ViT-H-14 (DFN)}& 99.00 & 24.30 & 22.33 & 27.23 & 53.03 \\
    & & \textit{ViT-SO400M-SigLIP} & 97.79 & 71.67 & 27.41 & 6.29 & 6.48 \\
    & & \textit{ViT-L-14 (datacomp)}& 98.89 & 16.51 & 21.29 & 26.92 & 48.52 \\
    & & \textit{xlm-roberta-large-ViT-H-14}& 99.46 & 17.15 & 16.63 & 20.18 & 35.64 \\
    & & \textit{ViT-L-14 (laion2b)} & 98.43 & 25.51 & 19.81 & 23.15 & 41.07 \\
    & & \textit{ViT-L-14 (openai)} & 89.79 & 26.33 & 20.74 & 24.69 & 50.29 \\
    & & \textit{ViT-B-32 (openai)} & 92.73 & 15.67 & 17.03 & 19.58 & 33.62 \\
    & & \textit{NegCLIP} & 96.83 & 15.50 & 17.54 & 22.58 & 36.40 \\

    \cmidrule{2-8}
    & \multirow{8}{*}{ComCO}    
    &  \textit{ViT-H-14 (DFN)}& 99.80 & 19.44 & 20.79 & 24.86 & 42.38 \\
    & & \textit{ViT-SO400M-SigLIP} & 97.63 & 70.57 & 32.34 & 5.42 & 5.72 \\
    & & \textit{ViT-L-14 (datacomp)}& 99.13 & 14.75 & 19.89 & 25.72 & 47.11 \\
    & & \textit{xlm-roberta-large-ViT-H-14}& 99.40 & 18.21 & 15.47 & 18.05 & 26.12 \\
    & & \textit{ViT-L-14 (laion2b)} & 98.76 & 20.91 & 18.11 & 20.77 & 33.54 \\
    & & \textit{ViT-L-14 (openai)} & 86.13 & 22.11 & 19.43 & 28.03 & 68.37 \\
    & & \textit{ViT-B-32 (openai)} & 91.20 & 15.56 & 13.31 & 19.66 & 39.39 \\
    & & \textit{NegCLIP} & 99.03 & 16.69 & 16.51 & 22.26 & 34.29 \\

    \bottomrule
    \end{tabular}
\end{table*}

\clearpage

\subsection{Text-based Object Retrieval}
\label{app:tor}

\subsubsection{Objective}
The Text-based Object Retrieval (TOR) experiment was designed to assess CLIP's text encoder's ability to retrieve individual objects from multi-object captions. This experiment aimed to investigate potential biases in object retrieval based on the object's position within the caption.

\begin{figure}[H]
    \centering
    \includegraphics[width=\linewidth]{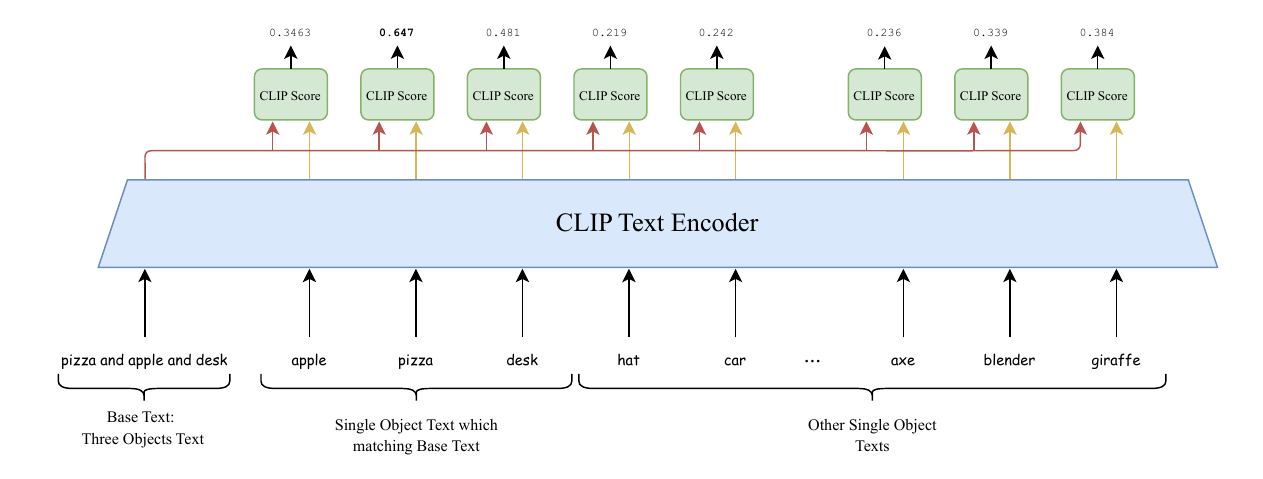}
    \caption{Visualization of the Text-based Object Retrieval experiment. This diagram illustrates the process of retrieving single-object texts based on multi-object captions using CLIP's text encoder. }
    \label{fig:tor_main_fig}
\end{figure}

\subsubsection{Methodology}

\begin{enumerate}
    \item \textbf{Dataset Preparation}: 
    \begin{itemize}
        \item We utilized both the SimCO and ComCO datasets, containing captions describing scenes with 2 to 5 objects.
        \item Each multi-object caption followed the format: ``Object1 and Object2 and ... and ObjectN''.
        \item We also prepared a set of single-object captions for each object class in our datasets.
    \end{itemize}

    \item \textbf{Text Embedding Generation}:
    \begin{itemize}
        \item We used CLIP's text encoder to generate embeddings for all multi-object captions.
        \item Similarly, we generated embeddings for all single-object captions.
    \end{itemize}

    \item \textbf{Similarity Computation}:
    \begin{itemize}
        \item For each multi-object caption, we computed the cosine similarity between its embedding and the embeddings of all single-object captions.
    \end{itemize}

    \item \textbf{Object Retrieval}:
    \begin{itemize}
        \item For each multi-object caption, we identified the single-object caption with the highest similarity score.
        \item We recorded which object from the multi-object caption (1st, 2nd, 3rd, etc.) matched this retrieved single-object caption.
    \end{itemize}

    \item \textbf{Evaluation}:
    \begin{itemize}
        \item We calculated the percentage of times each object position (1st, 2nd, 3rd, etc.) was retrieved as the most similar.
        \item This percentage represents the retrieval accuracy for each object position.
    \end{itemize}
\end{enumerate}

We repeated the TOR experiment on various models across scenarios with captions containing 2 to 5 objects. This was done to confirm the presence of the discovered bias. The complete results of this experiment, which was conducted on both the SIMCO and ComCO datasets, can be observed in Table \ref{tab:tor_total}.

\begin{table*}[htbp]
    \centering
    \scriptsize
    \setlength{\tabcolsep}{3pt}
    \renewcommand{\arraystretch}{1.1}
    \caption{Text-based Object Retrieval}
    \label{tab:tor_total}
    \begin{tabular}{llcccccc}
    \toprule
    \rowcolor[HTML]{E4E8F2}
    Number of Objects & Dataset & Model & \textbf{First Object} & \textbf{Second Object} & \textbf{Third Object} & \textbf{Fourth Object} & \textbf{Fifth Object} \\ 
    \midrule
    \multirow{16}{*}{n = 2}  & \multirow{8}{*}{SimCO}    
    &  \textit{ViT-H-14 (DFN)}& 69.18 & 30.82 & - & - & - \\
    & & \textit{ViT-SO400M-SigLIP} & 68.87 & 31.13 & - & - & - \\
    & & \textit{ViT-L-14 (datacomp)}& 69.93 & 30.07 & - & - & - \\
    & & \textit{xlm-roberta-large-ViT-H-14}& 78.95 & 21.05 & - & - & - \\
    & & \textit{ViT-L-14 (laion2b)} & 68.66 & 31.34 & - & - & - \\
    & & \textit{ViT-L-14 (openai)} & 75.82 & 24.18 & - & - & - \\
    & & \textit{ViT-B-32 (openai)} & 81.05 & 18.95 & - & - & - \\
    & & \textit{NegCLIP} & 77.78 & 22.22 & - & - & - \\

    \cmidrule{2-8}
    
    & \multirow{8}{*}{ComCO}    
    &  \textit{ViT-H-14 (DFN)}& 70.87 & 29.13 & - & - & - \\
    & & \textit{ViT-SO400M-SigLIP} & 67.56 & 32.44 & - & - & - \\
    & & \textit{ViT-L-14 (datacomp)}& 70.37 & 26.93 & - & - & - \\
    & & \textit{xlm-roberta-large-ViT-H-14}& 59.15 & 40.85 & - & - & - \\
    & & \textit{ViT-L-14 (laion2b)} & 70.84 & 29.16 & - & - & - \\
    & & \textit{ViT-L-14 (openai)} & 66.03 & 33.97 & - & - & - \\
    & & \textit{ViT-B-32 (openai)} & 61.62  & 38.38 & - & - & - \\
    & & \textit{NegCLIP} & 64.13 & 35.87 & - & - & - \\
    
    \midrule 
    \multirow{16}{*}{n = 3} & \multirow{8}{*}{SimCO}   
    &  \textit{ViT-H-14 (DFN)}& 62.05 & 18.07 & 19.88 & - & - \\
    & & \textit{ViT-SO400M-SigLIP} & 58.05 & 20.50 & 21.46 & - & - \\
    & & \textit{ViT-L-14 (datacomp)}& 61.68 & 20.35 & 17.96 & - & - \\
    & & \textit{xlm-roberta-large-ViT-H-14}& 66.75 & 23.86 & 9.39 & - & - \\
    & & \textit{ViT-L-14 (laion2b)} & 62.31 & 12.56 & 25.13 & - & - \\
    & & \textit{ViT-L-14 (openai)} & 65.71 & 16.67 & 17.62 & - & - \\
    & & \textit{ViT-B-32 (openai)} & 74.23 & 13.62 & 12.15 & - & - \\
    & & \textit{NegCLIP} & 77.43 & 13.75 & 8.83 & - & - \\

    \cmidrule{2-8}
    & \multirow{8}{*}{ComCO}    
    &  \textit{ViT-H-14 (DFN)}& 67.08 & 22.19 & 10.73 & - & - \\
    & & \textit{ViT-SO400M-SigLIP} & 61.11 & 23.33 & 15.56 & - & - \\
    & & \textit{ViT-L-14 (datacomp)}& 72.23 & 19.05 & 8.72 & - & - \\
    & & \textit{xlm-roberta-large-ViT-H-14}& 43.60 & 31.36 & 25.05 & - & - \\
    & & \textit{ViT-L-14 (laion2b)} & 66.85 & 23.52 & 9.63 & - & - \\
    & & \textit{ViT-L-14 (openai)} & 57.66 & 26.75 & 15.59 & - & - \\
    & & \textit{ViT-B-32 (openai)} & 55.73 & 28.28 & 15.98 & - & - \\
    & & \textit{NegCLIP} & 57.56 & 29.45 & 12.99 & - & - \\
     
    \midrule
    \multirow{16}{*}{n = 4} & \multirow{8}{*}{SimCO}   
    
    &  \textit{ViT-H-14 (DFN)}& 60.06 & 12.77 & 12.03 & 15.14 & - \\
    & & \textit{ViT-SO400M-SigLIP} & 53.54 & 14.76 & 11.43 & 20.27 & - \\
    & & \textit{ViT-L-14 (datacomp)}& 62.16 & 15.99 & 10.41 & 11.44 & - \\
    & & \textit{xlm-roberta-large-ViT-H-14}& 62.58 & 22.52 & 10.91 & 3.99 & - \\
    & & \textit{ViT-L-14 (laion2b)} & 67.81 & 8.97 & 5.80 & 17.41 & - \\
    & & \textit{ViT-L-14 (openai)} & 66.87 & 11.59 & 6.18 & 15.35 & - \\
    & & \textit{ViT-B-32 (openai)} & 76.37 & 10.03 & 7.50 & 6.55 & - \\
    & & \textit{NegCLIP} & 82.90 & 10.20 & 4.61 & 2.29 & - \\

    \cmidrule{2-8}
    & \multirow{8}{*}{ComCO}    
    &  \textit{ViT-H-14 (DFN)}& 64.34 & 19.25 & 11.14 & 5.27 & - \\
    & & \textit{ViT-SO400M-SigLIP} & 58.11 & 21.16 & 10.99 & 9.73 & - \\
    & & \textit{ViT-L-14 (datacomp)}& 71.13 & 16.26 & 8.74 & 3.87 & - \\
    & & \textit{xlm-roberta-large-ViT-H-14}& 44.03 & 23.73 & 18.07 & 14.18 & - \\
    & & \textit{ViT-L-14 (laion2b)} & 63.96 & 21.59 & 10.68 & 3.76 & - \\
    & & \textit{ViT-L-14 (openai)} & 48.20 & 26.01 & 10.74 & 8.74 & - \\
    & & \textit{ViT-B-32 (openai)} & 50.31 & 20.74 & 15.45 & 6.79 & - \\
    & & \textit{NegCLIP} & 51.63 & 28.92 & 14.86 & 4.59 & - \\
     
    \midrule
    \multirow{16}{*}{n = 5}& \multirow{8}{*}{SimCO}

    &  \textit{ViT-H-14 (DFN)}& 60.80 & 10.61 & 8.35 & 9.02 & 11.22 \\
    & & \textit{ViT-SO400M-SigLIP} & 49.47 & 13.32 & 3.39 & 11.97 & 21.25 \\
    & & \textit{ViT-L-14 (datacomp)}& 66.43 & 16.12 & 6.59 & 4.99 & 5.87 \\
    & & \textit{xlm-roberta-large-ViT-H-14}& 60.65 & 21.03 & 11.90 & 5.15 & 1.28 \\
    & & \textit{ViT-L-14 (laion2b)} & 74.07 & 9.51 & 4.48 & 2.80 & 9.14 \\
    & & \textit{ViT-L-14 (openai)} & 71.71 & 10.59 & 2.99 & 2.71 & 12.00 \\
    & & \textit{ViT-B-32 (openai)} & 43.86 & 26.41 & 15.44 & 8.57 & 5.72 \\
    & & \textit{NegCLIP} & 85.00 & 10.39 & 3.12 &1.24 & 0.26 \\

    \cmidrule{2-8}
    & \multirow{8}{*}{ComCO}    
    &  \textit{ViT-H-14 (DFN)}& 61.06 & 17.00 & 11.98 & 6.69 & 3.27 \\
    & & \textit{ViT-SO400M-SigLIP} & 55.77 & 19.25 & 10.24 & 6.73 & 8.01 \\
    & & \textit{ViT-L-14 (datacomp)}& 68.96 & 14.61 & 9.40 & 4.77 & 2.25 \\
    & & \textit{xlm-roberta-large-ViT-H-14}& 28.86 & 26.87 & 19.42 & 14.61 & 10.24 \\
    & & \textit{ViT-L-14 (laion2b)} & 61.93 & 19.10 & 11.65 & 5.11 & 2.21 \\
    & & \textit{ViT-L-14 (openai)} & 38.40 & 24.80 & 18.79 & 11.04 & 6.68 \\
    & & \textit{ViT-B-32 (openai)} & 44.71 & 26.69 & 16.44 & 8.37 & 3.79 \\
    & & \textit{NegCLIP} & 45.70 & 27.56 & 17.03 & 7.57 & 2.15 \\

    \bottomrule
    \end{tabular}
    \end{table*}

\clearpage
\subsection{Image-based Object Classification}
\label{app:ioc}

\subsubsection{Objective}
The Image-based Object Classification (IOC) experiment was designed to evaluate CLIP's image encoder's ability to represent individual objects within multi-object images. This experiment aimed to investigate potential biases in object classification based on the object's size within the image.

\begin{figure}[H]
    \centering
    \includegraphics[width=0.8\linewidth]{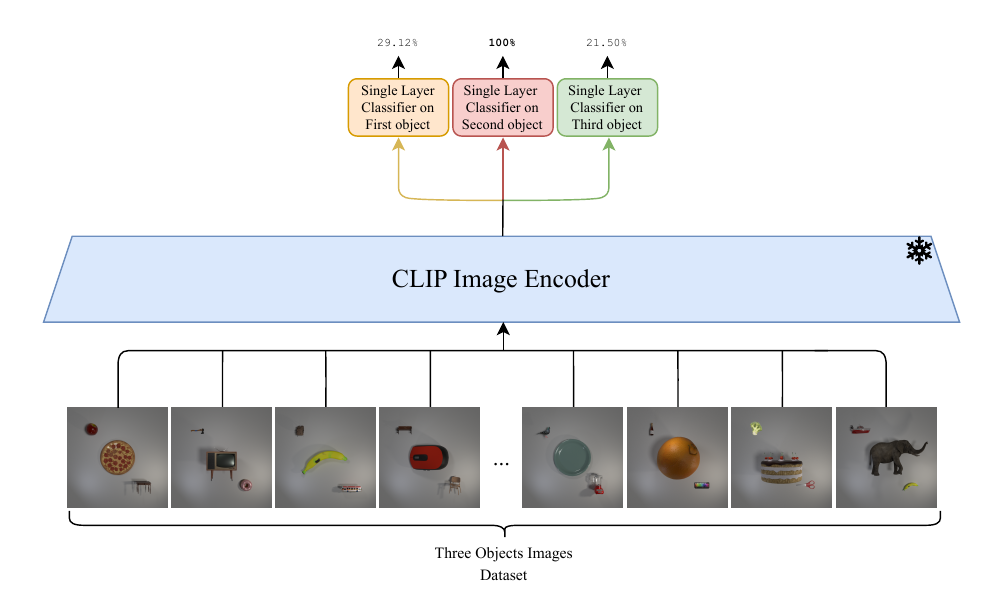}
    \caption{ Illustration of the Image-based Object Classification experiment with the ComCO dataset. The diagram shows the process of classifying individual objects in K-object images using CLIP's image encoder, with a single-layer classifier trained on the generated image embeddings}

    \label{fig:ioc_main_fig}
\end{figure}
\subsubsection{Methodology}

\begin{enumerate}
    \item \textbf{Dataset Preparation}: 
    \begin{itemize}
        \item We utilized both the SimCO and ComCO datasets, containing images with 2 to 5 objects.
        \item In each image, one object was deliberately made larger than the others.
        \item The position of the larger object was varied across images to avoid position-based biases.
    \end{itemize}

    \item \textbf{Image Embedding Generation}:
    \begin{itemize}
        \item For each multi-object image, we used CLIP's image encoder to generate an image embedding.
        \item This embedding is a high-dimensional vector representation of the entire image.
    \end{itemize}

    \item \textbf{Classifier Training}:
    \begin{itemize}
        \item We trained separate single-layer classifiers for each object position (large object, small object 1, small object 2, etc.).
        \item Input: The image embedding of the multi-object image.
        \item Output: The predicted object class for that specific position/size.
    \end{itemize}

    \item \textbf{Evaluation}:
    \begin{itemize}
        \item We tested each classifier on a held-out portion of the dataset.
        \item For each image, we recorded whether the classifier correctly identified the object at its respective position/size.
        \item We calculated the classification accuracy for each object position/size across all test images.
    \end{itemize}
\end{enumerate}

We conducted the IOC experiment on images from both datasets, focusing on scenarios with one significantly larger object in varying positions. The experiment was repeated across models, and the average results are shown in Table \ref{tab:ioc_total}.

\begin{table*}[htbp]
    \centering
    \scriptsize
    \setlength{\tabcolsep}{4pt}
    \renewcommand{\arraystretch}{1.1}
    \caption{Image-based Object Classification}
    \label{tab:ioc_total}
    \begin{tabular}{lllccccc}
    \toprule
    \rowcolor[HTML]{E4E8F2}
    Number of Objects & Dataset & Model & \textbf{Large Object} & \textbf{Small Obj 1} & \textbf{Small Obj 2} & \textbf{Small Obj 3} & \textbf{Small Obj 4} \\ 
    \midrule
    \multirow{16}{*}{n = 2}  & \multirow{8}{*}{SimCO}    
    &  \textit{ViT-H-14 (DFN)}& 88.1 & 14.29 & - & - & - \\
    & & \textit{ViT-SO400M-SigLIP} & 97.62 & 16.67 & - & - & - \\
    & & \textit{ViT-L-14 (datacomp)}& 83.33 & 11.9 & - & - & -  \\
    & & \textit{xlm-roberta-large-ViT-H-14}& 78.57 & 21.43 & - & - & -  \\
    & & \textit{ViT-L-14 (laion2b)} & 66.67 & 11.9 & - & - & -  \\
    & & \textit{ViT-L-14 (openai)} & 64.29 & 0.00 & - & - & -  \\
    & & \textit{ViT-B-32 (openai)} & 61.9 & 0.00 & - & - & -  \\
    & & \textit{NegCLIP} & 40.48 & 7.14 & - & - & - \\

    \cmidrule{2-8}
    
    & \multirow{8}{*}{ComCO}    
    &  \textit{ViT-H-14 (DFN)}& 100.0 & 26.36 & - & - & - \\
    & & \textit{ViT-SO400M-SigLIP} & 100.0 & 33.9 & - & - & - \\
    & & \textit{ViT-L-14 (datacomp)}& 100.0 & 42.35 & - & - & - \\
    & & \textit{xlm-roberta-large-ViT-H-14}& 100.0 & 40.85 & - & - & - \\
    & & \textit{ViT-L-14 (laion2b)} & 100.0 & 31.29 & - & - & - \\
    & & \textit{ViT-L-14 (openai)} & 99.8 & 41.29 & - & - & - \\
    & & \textit{ViT-B-32 (openai)} & 99.8 & 35.81 & - & - & - \\
    & & \textit{NegCLIP} & 99.6 & 41.95 & - & - & - \\
    
    \midrule 
    \multirow{16}{*}{n = 3} & \multirow{8}{*}{SimCO}   
    &  \textit{ViT-H-14 (DFN)}& 100.0 & 35.65 & 41.57 & - & -  \\
    & & \textit{ViT-SO400M-SigLIP} & 99.8 & 42.8 & 49.03 & - & -  \\
    & & \textit{ViT-L-14 (datacomp)}& 100.0 & 39.94 & 51.28 & - & -  \\
    & & \textit{xlm-roberta-large-ViT-H-14}& 99.9 & 48.42 & 56.28 & - & -  \\
    & & \textit{ViT-L-14 (laion2b)} & 99.8 & 45.56 & 56.08 & - & -  \\
    & & \textit{ViT-L-14 (openai)} & 98.98 & 39.73 & 50.46 & - & -  \\
    & & \textit{ViT-B-32 (openai)} & 96.12 & 38.1 & 51.58 & - & -  \\
    & & \textit{NegCLIP} & 97.04 & 42.59 & 59.35 & - & -  \\

    \cmidrule{2-8}
    & \multirow{8}{*}{ComCO}    
    &  \textit{ViT-H-14 (DFN)}& 100.0 & 29.12 & 21.5 & - & - \\
    & & \textit{ViT-SO400M-SigLIP} & 100.0 & 30.94 & 29.94 & - & - \\
    & & \textit{ViT-L-14 (datacomp)}& 100.0 & 36.56 & 33.5 & - & - \\
    & & \textit{xlm-roberta-large-ViT-H-14}& 100.0 & 33.69 & 32.31 & - & - \\
    & & \textit{ViT-L-14 (laion2b)} & 100.0 & 35.44 & 30.31 & - & - \\
    & & \textit{ViT-L-14 (openai)} & 99.94 & 33.31 & 34.31 & - & - \\
    & & \textit{ViT-B-32 (openai)} & 99.94 & 29.0 & 32.94 & - & - \\
    & & \textit{NegCLIP} & 99.81 & 33.88 & 43.0 & - & - \\
     
    \midrule
    \multirow{16}{*}{n = 4} & \multirow{8}{*}{SimCO}   
    
    &  \textit{ViT-H-14 (DFN)}& 100.0 & 40.06 & 34.06 & 41.31 & -  \\
    & & \textit{ViT-SO400M-SigLIP} & 100.0 & 47.0 & 38.5 & 41.06 & -  \\
    & & \textit{ViT-L-14 (datacomp)}& 100.0 & 48.94 & 38.38 & 45.06 & - \\
    & & \textit{xlm-roberta-large-ViT-H-14}& 100.0 & 48.19 & 35.81 & 46.38 & -  \\
    & & \textit{ViT-L-14 (laion2b)} & 100.0 & 50.5 & 41.81 & 43.94 & - \\
    & & \textit{ViT-L-14 (openai)} & 100.0 & 45.19 & 38.38 & 39.0  & - \\
    & & \textit{ViT-B-32 (openai)} & 100.0 & 38.06 & 31.5 & 37.25  & - \\
    & & \textit{NegCLIP} & 100.0 & 42.0 & 37.25 & 46.94  & - \\

    \cmidrule{2-8}
    & \multirow{8}{*}{ComCO}    
    &  \textit{ViT-H-14 (DFN)}& 100.0 & 16.64 & 14.13 & 12.38 & - \\
    & & \textit{ViT-SO400M-SigLIP} & 100.0 & 18.95 & 15.57 & 17.57 & - \\
    & & \textit{ViT-L-14 (datacomp)}& 100.0 & 20.64 & 21.01 & 19.01 & - \\
    & & \textit{xlm-roberta-large-ViT-H-14}& 100.0 & 20.45 & 18.45 & 16.51 & - \\
    & & \textit{ViT-L-14 (laion2b)} & 100.0 & 19.76 & 17.57 & 18.89 & - \\
    & & \textit{ViT-L-14 (openai)} & 99.94 & 19.32 & 21.89 & 22.39 & - \\
    & & \textit{ViT-B-32 (openai)} & 100.0 & 21.58 & 21.83 & 22.26 & - \\
    & & \textit{NegCLIP} & 100.0 & 21.89 & 23.64 & 31.33 & - \\
     
    \midrule
    \multirow{16}{*}{n = 5}& \multirow{8}{*}{SimCO}

    &  \textit{ViT-H-14 (DFN)}& 100.0 & 34.0 & 30.0 & 30.38 & 21.62 \\
    & & \textit{ViT-SO400M-SigLIP} & 100.0 & 38.5 & 34.7 & 27.38 & 25.62 \\
    & & \textit{ViT-L-14 (datacomp)}& 100.0 & 40.38 & 36.12 & 32.0 & 24.75 \\
    & & \textit{xlm-roberta-large-ViT-H-14}& 100.0 & 41.56 & 39.56 & 36.69 & 32.81 \\
    & & \textit{ViT-L-14 (laion2b)} & 100.0 & 43.88 & 39.5 & 34.0 & 28.94 \\
    & & \textit{ViT-L-14 (openai)} & 100.0 & 42.19 & 36.38 & 32.81 & 31.94 \\
    & & \textit{ViT-B-32 (openai)} & 98.81 & 36.25 & 35.38 & 33.88 & 26.06 \\
    & & \textit{NegCLIP} & 99.19 & 40.88 & 37.94 & 37.56 & 28.94 \\

    \cmidrule{2-8}
    & \multirow{8}{*}{ComCO}    
    &  \textit{ViT-H-14 (DFN)}& 100.0 & 13.88 & 9.38 & 9.32 & 11.94 \\
    & & \textit{ViT-SO400M-SigLIP} & 100.0 & 15.51 & 13.88 & 14.57 & 14.76 \\
    & & \textit{ViT-L-14 (datacomp)}& 100.0 & 18.2 & 15.07 & 16.07 & 18.32 \\
    & & \textit{xlm-roberta-large-ViT-H-14}& 99.94 & 15.38 & 14.88 & 15.26 & 19.14 \\
    & & \textit{ViT-L-14 (laion2b)} & 100.0 & 15.51 & 12.32 & 14.13 & 17.95 \\
    & & \textit{ViT-L-14 (openai)} & 100.0 & 15.38 & 14.76 & 16.76 & 20.01 \\
    & & \textit{ViT-B-32 (openai)} & 99.87 & 17.76 & 18.64 & 19.2 & 23.14 \\
    & & \textit{NegCLIP} & 100 & 18.89 & 16.57 & 23.51 & 28.77 \\

    \bottomrule
    \end{tabular}
\end{table*}

\clearpage

\subsection{Image-based Object Retrieval}
\label{app:ior}
\subsubsection{Objective}
The Image-based Object Retrieval (IOR) experiment was designed to assess CLIP's image encoder's ability to retrieve individual objects from multi-object images. This experiment aimed to investigate potential biases in object retrieval based on the object's size within the image.

\begin{figure}[H]
    \centering
    \includegraphics[width=0.8\linewidth]{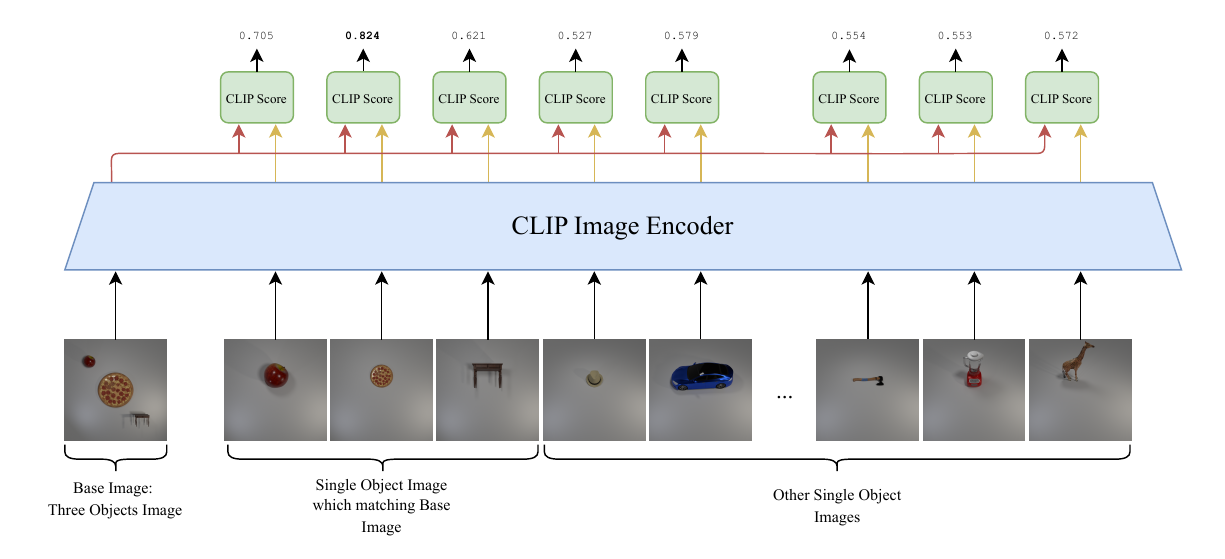}
    \caption{Visualization of the Image-based Object Retrieval experiment. This diagram illustrates the process of retrieving single-object images based on multi-object image inputs using CLIP's image encoder. The experiment employs a base image containing three objects of varying sizes. CLIP scores are computed between the embedding of this multi-object image and embeddings of various single-object images.}

    \label{fig:ior_main_fig}
\end{figure}

\subsubsection{Methodology}

\begin{enumerate}
    \item \textbf{Dataset Preparation}: 
    \begin{itemize}
        \item We utilized both the SimCO and ComCO datasets, containing images with 2 to 5 objects.
        \item In each multi-object image, one object was deliberately made larger than the others.
        \item The position of the larger object was varied across images to avoid position-based biases.
        \item We also prepared a set of single-object images for each object class in our datasets.
    \end{itemize}

    \item \textbf{Image Embedding Generation}:
    \begin{itemize}
        \item We used CLIP's image encoder to generate embeddings for all multi-object images.
        \item Similarly, we generated embeddings for all single-object images.
    \end{itemize}

    \item \textbf{Similarity Computation}:
    \begin{itemize}
        \item For each multi-object image, we computed the cosine similarity between its embedding and the embeddings of all single-object images.
    \end{itemize}

    \item \textbf{Object Retrieval}:
    \begin{itemize}
        \item For each multi-object image, we identified the single-object image with the highest similarity score.
        \item We recorded whether the retrieved single-object image corresponded to the large object or one of the small objects in the multi-object image.
    \end{itemize}

    \item \textbf{Evaluation}:
    \begin{itemize}
        \item We calculated the percentage of times the large object and each small object were retrieved as the most similar.
        \item This percentage represents the retrieval accuracy for each object size category (large object, small object 1, small object 2, etc.).
    \end{itemize}
\end{enumerate}

We conducted the IOR experiment on images from the SimCO and ComCO datasets with 2 to 5 objects, varying the position of the larger object to avoid location-based biases. The results are shown in Table \ref{tab:ior_total}.
\begin{table*}[htbp]
    \centering
    \scriptsize
    \setlength{\tabcolsep}{3pt}
    \renewcommand{\arraystretch}{1.1}
    \caption{Image-based Object Retrieval}
    \label{tab:ior_total}
    \begin{tabular}{llcccccc}
    \toprule
    \rowcolor[HTML]{E4E8F2}
    Number of Objects & Dataset & Model & \textbf{Large Object} & \textbf{Small Obj 1} & \textbf{Small Obj 2} & \textbf{Small Obj 3} & \textbf{Small Obj 4} \\ 
    \midrule
    \multirow{16}{*}{n = 2}  & \multirow{8}{*}{SimCO}    
    &  \textit{ViT-H-14 (DFN)}& 99.11 & 0.89 & - & - & - \\
    & & \textit{ViT-SO400M-SigLIP}& 91.67 & 8.33 & - & - & -  \\
    & & \textit{ViT-L-14 (datacomp)}& 91.96 & 8.04 & - & - & - \\
    & & \textit{xlm-roberta-large-ViT-H-14}& 94.92 & 5.08 & - & - & - \\
    & & \textit{ViT-L-14 (laion2b)}& 92.86 & 7.14 & - & - & - \\
    & & \textit{ViT-L-14 (openai)}& 87.88 & 12.12 & - & - & - \\
    & & \textit{ViT-B-32 (openai)}& 90.24 & 9.76 & - & - & - \\
    & & \textit{NegCLIP}& 94.64 & 5.36 & - & - & - \\

    \cmidrule{2-8}
    
    & \multirow{8}{*}{ComCO}    
    &  \textit{ViT-H-14 (DFN)}& 97.35 & 2.65 & - & - & - \\
    & & \textit{ViT-SO400M-SigLIP}& 95.13 & 4.87 & - & - & - \\
    & & \textit{ViT-L-14 (datacomp)} & 89.85 & 10.15 & - & - & - \\
    & & \textit{xlm-roberta-large-ViT-H-14}& 93.89 & 6.11 & - & - & - \\
    & & \textit{ViT-L-14 (laion2b)}& 94.84 & 5.16 & - & - & - \\
    & & \textit{ViT-L-14 (openai)} & 83.7 & 16.30 & - & - & -\\
    & & \textit{ViT-B-32 (openai)}& 86.86 & 13.14 & - & - & - \\
    & & \textit{NegCLIP}& 83.3 & 16.7 & - & - & - \\
    
    \midrule 
    \multirow{16}{*}{n = 3} & \multirow{8}{*}{SimCO}   
    &  \textit{ViT-H-14 (DFN)}& 93.80 & 0.65 & 5.55 & - & - \\
    & & \textit{ViT-SO400M-SigLIP}& 83.27 & 5.61 & 11.12 & - & - \\
    & & \textit{ViT-L-14 (datacomp)}& 77.16 & 5.81 & 17.04 & - & - \\
    & & \textit{xlm-roberta-large-ViT-H-14} & 80.21 & 5.12 & 14.66 & - & -\\
    & & \textit{ViT-L-14 (laion2b)}& 76.57 & 9.57 & 13.86 & - & - \\
    & & \textit{ViT-L-14 (openai)}& 72.07 & 8.66 & 19.27 & - & - \\
    & & \textit{ViT-B-32 (openai)}& 61.14 & 14.69 & 24.17 & - & - \\
    & & \textit{NegCLIP}& 59.13 & 14.91 & 25.96 & - & - \\

    \cmidrule{2-8}
    & \multirow{8}{*}{ComCO}    
    &  \textit{ViT-H-14 (DFN)}& 96.52 & 1.71 & 17.8 & - & - \\
    & & \textit{ViT-SO400M-SigLIP}& 90.5 & 5.47 & 4.03 & - & - \\
    & & \textit{ViT-L-14 (datacomp)}& 89.65 & 6.09 & 4.26 & - & - \\
    & & \textit{xlm-roberta-large-ViT-H-14}& 91.39 & 4.92 & 3.69 & - & - \\
    & & \textit{ViT-L-14 (laion2b)}& 91.26 & 3.28 & 5.46 & - & - \\
    & & \textit{ViT-L-14 (openai)}& 74.2 & 12.79 & 13.01 & - & - \\
    & & \textit{ViT-B-32 (openai)}& 80.6 & 5.22 & 14.18 & - & - \\
    & & \textit{NegCLIP}& 76.36 & 10.47 & 13.18 & - & - \\
     
    \midrule
    \multirow{16}{*}{n = 4} & \multirow{8}{*}{SimCO}   
    
    &  \textit{ViT-H-14 (DFN)}& 99.5 & 0.0 & 0.0 & 0.5 & - \\
    & & \textit{ViT-SO400M-SigLIP}& 91.03 & 1.28 & 2.99 & 4.7 & - \\
    & & \textit{ViT-L-14 (datacomp)} & 89.71 & 3.43 & 3.61 & 3.25 & -\\
    & & \textit{xlm-roberta-large-ViT-H-14}& 92.47 & 2.08 & 2.60 & 2.86 & - \\
    & & \textit{ViT-L-14 (laion2b)} & 86.92 & 4.67 & 3.74 & 4.67 & -\\
    & & \textit{ViT-L-14 (openai)}& 70.55 & 13.01 & 7.53 & 8.9 & - \\
    & & \textit{ViT-B-32 (openai)}& 52.17 & 18.84 & 13.04 & 15.94 & - \\
    & & \textit{NegCLIP}& 74.4 & 10.4 & 7.2 & 8.0 & - \\

    \cmidrule{2-8}
    & \multirow{8}{*}{ComCO}    
    &  \textit{ViT-H-14 (DFN)}& 95.86 & 2.55 & 1.27 & 0.32 & - \\
    & & \textit{ViT-SO400M-SigLIP}& 94.03 & 2.24 & 1.49 & 2.24 & - \\
    & & \textit{ViT-L-14 (datacomp)}& 93.3 & 3.91 & 1.12 & 16.8 & - \\
    & & \textit{xlm-roberta-large-ViT-H-14} & 90.91 & 2.02 & 5.05 & 2.02 & -\\
    & & \textit{ViT-L-14 (laion2b)} & 91.78 & 5.48 & 2.74 & 0.0 & -\\
    & & \textit{ViT-L-14 (openai)}& 67.86 & 14.29 & 7.14 & 10.71 & - \\
    & & \textit{ViT-B-32 (openai)}& 85.0 & 0.0 & 5.0 & 10.0 & - \\
    & & \textit{NegCLIP}& 79.55 & 0.0 & 2.27 & 18.19 & - \\
     
    \midrule
    \multirow{16}{*}{n = 5}& \multirow{8}{*}{SimCO}

    &  \textit{ViT-H-14 (DFN)}& 100.0 & 0.0 & 0.0 & 0.0 & 0.0 \\
    & & \textit{ViT-SO400M-SigLIP}& 94.92 & 3.39 & 1.69 & 0.0 & 0.0 \\
    & & \textit{ViT-L-14 (datacomp)}& 91.3 & 5.59 & 1.24 & 1.24 & 0.62 \\
    & & \textit{xlm-roberta-large-ViT-H-14}& 77.42 & 11.83 & 5.38 & 3.23 & 2.15 \\
    & & \textit{ViT-L-14 (laion2b)}& 81.01 & 8.86 & 5.06 & 1.27 & 0.38 \\
    & & \textit{ViT-L-14 (openai)}& 77.14 & 8.57 & 5.71 & 5.71 & 2.86 \\
    & & \textit{ViT-B-32 (openai)}& 68.75 & 25.0 & 6.25 & 0.0 & 0.0 \\
    & & \textit{NegCLIP}& 58.62 & 17.24 & 15.52 & 5.17 & 3.45 \\

    \cmidrule{2-8}
    & \multirow{8}{*}{ComCO}    
    &  \textit{ViT-H-14 (DFN)}& 95.16 & 1.61  & 1.61 & 0.0 & 1.61 \\
    & & \textit{ViT-SO400M-SigLIP}& 80.0 & 0.0 & 0.0 & 0.0 & 20.0 \\
    & & \textit{ViT-L-14 (datacomp)}& 90.91 & 4.55 & 0.0 & 0.0 & 4.55 \\
    & & \textit{xlm-roberta-large-ViT-H-14}& 100.0 & 0.0 & 0.0 & 0.0 & 0.0 \\
    & & \textit{ViT-L-14 (laion2b)}& 100.0 & 0.0 & 0.0 & 0.0 & 0.0 \\
    & & \textit{ViT-L-14 (openai)} & 100.0 & 0.0 & 0.0 & 0.0 & 0.0 \\
    & & \textit{ViT-B-32 (openai)}& 100.0 & 0.0 & 0.0 & 0.0 & 0. \\
    & & \textit{NegCLIP}& 50.0 & 0.0 & 0.0 & 50.0 & 0.0 \\

    \bottomrule
    \end{tabular}
\end{table*}

\clearpage 
\subsection{Text-based Object Classification for Long Caption}
\label{app:toc-long}

In this section, we revisited the IOC experiment with a significant modification to the caption structure. Our objective was to investigate whether the previously observed bias persists in longer, more elaborate captions. We achieved this by expanding the caption template, incorporating additional descriptive phrases between object mentions.

The extended caption template used in this experiment was as follows:


\begin{figure}[htbp]
    \centering
    \includegraphics[width=0.9\linewidth]{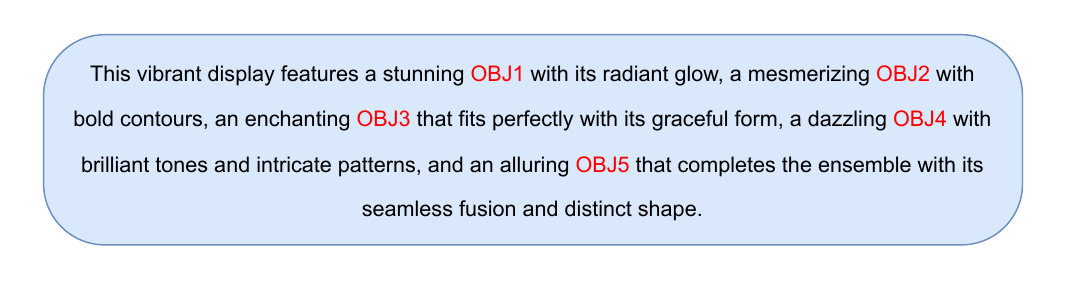}
    \caption{Format for Extended Caption Template}
    \label{fig:comco}
\end{figure}

This template allowed us to maintain a consistent structure while significantly increasing the caption length and complexity.

The results of this modified IOC experiment are presented in Table \ref{tab:toc_long_total}. Notably, the observed pattern closely resembles that of the standard IOC experiment. This similarity suggests that the bias identified in shorter captions persists even in more elaborate textual descriptions.

\begin{table*}[ht]
    \centering
    \scriptsize
    \setlength{\tabcolsep}{4pt}
    \renewcommand{\arraystretch}{1.1}
    \caption{Text-based Object Classification on Long Captions}
    \label{tab:toc_long_total}
    \begin{tabular}{lllccccc}
    \toprule
    \rowcolor[HTML]{E4E8F2}
    Number of Objects & Dataset & Model & \textbf{First Object} & \textbf{Second Object} & \textbf{Third Object} & \textbf{Fourth Object} & \textbf{Fifth Object} \\ 
    \midrule
    \multirow{16}{*}{n = 2}  & \multirow{8}{*}{SimCO}    
    &  \textit{ViT-H-14 (DFN)}& 100.0 & 89.01 & - & - & - \\
    & & \textit{ViT-SO400M-SigLIP} & 100.0 & 93.83 & - & - & - \\
    & & \textit{ViT-L-14 (datacomp)}& 100.0 & 63.22 & - & - & - \\
    & & \textit{xlm-roberta-large-ViT-H-14}& 99.82 & 51.83 & - & - & - \\
    & & \textit{ViT-L-14 (laion2b)} & 100.0 & 85.88 & - & - & - \\
    & & \textit{ViT-L-14 (openai)} & 99.65 & 98.26 & - & - & - \\
    & & \textit{ViT-B-32 (openai)} & 100.0 & 72.69 & - & - & - \\
    & & \textit{NegCLIP} & 100 & 89.59  & - & - & - \\

    \cmidrule{2-8}
    
    & \multirow{8}{*}{ComCO}    
    &  \textit{ViT-H-14 (DFN)}& 99.99 & 99.86 & - & - & - \\
    & & \textit{ViT-SO400M-SigLIP} & 100 & 99.48 & - & - & - \\
    & & \textit{ViT-L-14 (datacomp)}& 100 & 98.89 & - & - & - \\
    & & \textit{xlm-roberta-large-ViT-H-14}& 99.95 & 92.84 & - & - & - \\
    & & \textit{ViT-L-14 (laion2b)} & 100 & 99.03 & - & - & - \\
    & & \textit{ViT-L-14 (openai)} & 99.99 & 99.99 & - & - & - \\
    & & \textit{ViT-B-32 (openai)} & 99.59 & 99.45 & - & - & - \\
    & & \textit{NegCLIP} & 99.94 & 98.99 & - & - & - \\
    
    \midrule 
    \multirow{16}{*}{n = 3} & \multirow{8}{*}{SimCO}   
    &  \textit{ViT-H-14 (DFN)}& 99.34 & 43.49 & 89.66 & - & - \\
    & & \textit{ViT-SO400M-SigLIP} & 100.0 & 65.26 & 49.76 & - & - \\
    & & \textit{ViT-L-14 (datacomp)}& 100.0 & 30.47 & 37.20 & - & - \\
    & & \textit{xlm-roberta-large-ViT-H-14}& 97.78 & 22.96 & 27.23 & - & - \\
    & & \textit{ViT-L-14 (laion2b)} & 99.65 & 57.67 & 35.51 & - & - \\
    & & \textit{ViT-L-14 (openai)} & 99.13 & 86.67  & 58.22 & - & - \\
    & & \textit{ViT-B-32 (openai)} & 96.26 & 54.19 & 44.88 & - & - \\
    & & \textit{NegCLIP} & 98.30 & 67.60 & 65.90 & - & - \\

    \cmidrule{2-8}
    & \multirow{8}{*}{ComCO}    
    &  \textit{ViT-H-14 (DFN)}& 99.31 & 78.44 & 84.15 & - & - \\
    & & \textit{ViT-SO400M-SigLIP} & 99.93 & 67.22 & 76.89 & - & - \\
    & & \textit{ViT-L-14 (datacomp)}& 98.98 & 85.77 & 65.64 & - & - \\
    & & \textit{xlm-roberta-large-ViT-H-14}& 99.21 & 38.60 & 60.10 & - & - \\
    & & \textit{ViT-L-14 (laion2b)} & 98.81 & 82.72 & 74.31 & - & - \\
    & & \textit{ViT-L-14 (openai)} & 99.41 & 96.44 & 82.18 & - & - \\
    & & \textit{ViT-B-32 (openai)} & 95.59 & 81.91 & 76.09 & - & - \\
    & & \textit{NegCLIP} & 98.62 & 74.29 & 81.70 & - & - \\
     
    \midrule
    \multirow{16}{*}{n = 4} & \multirow{8}{*}{SimCO}   
    
    &  \textit{ViT-H-14 (DFN)}& 99.17 & 24.74 & 67.00 & 41.46 & - \\
    & & \textit{ViT-SO400M-SigLIP} & 100.0 & 46.75 & 24.40 & 20.93 & - \\
    & & \textit{ViT-L-14 (datacomp)}& 100.0 & 15.27 & 17.79 & 43.03 & - \\
    & & \textit{xlm-roberta-large-ViT-H-14}& 98.87 & 13.34 & 12.67 & 15.85 & - \\
    & & \textit{ViT-L-14 (laion2b)} & 99.56 & 36.03 & 19.23 & 34.51 & - \\
    & & \textit{ViT-L-14 (openai)} & 98.22 & 70.29 & 40.54 & 50.71 & - \\
    & & \textit{ViT-B-32 (openai)} & 97.47 & 41.20 & 25.18 & 24.31 & - \\
    & & \textit{NegCLIP} & 98.93 & 49.58 & 35.89 & 35.40 & - \\

    \cmidrule{2-8}
    & \multirow{8}{*}{ComCO}    
    &  \textit{ViT-H-14 (DFN)}& 98.34 & 62.49 & 70.25 & 42.34 & - \\
    & & \textit{ViT-SO400M-SigLIP} & 99.90 & 39.28 & 58.01 & 32.51 & - \\
    & & \textit{ViT-L-14 (datacomp)}& 97.95 & 71.61 & 37.24 & 48.50 & - \\
    & & \textit{xlm-roberta-large-ViT-H-14} & 99.34 & 20.38 & 21.45 & 25.08 & - \\
    & & \textit{ViT-L-14 (laion2b)}& 98.41 & 66.90 & 51.43 & 38.87 & - \\
    & & \textit{ViT-L-14 (openai)} & 96.39 & 88.74 & 62.87 & 75.1 & - \\
    & & \textit{ViT-B-32 (openai)} & 96.81 & 62.50 & 59.19 & 22.93 & - \\
    & & \textit{NegCLIP} & 98.50 & 45.93 & 40.11 & 68.58 & - \\
     
    \midrule
    \multirow{16}{*}{n = 5}& \multirow{8}{*}{SimCO}

    &  \textit{ViT-H-14 (DFN)}& 97.44 & 18.82 & 53.68  & 26.08 & 47.45\\
    & & \textit{ViT-SO400M-SigLIP} & 100.0 & 20.35 & 19.30 & 12.57 & 18.40 \\
    & & \textit{ViT-L-14 (datacomp)}& 99.74 & 17.57 & 19.29 & 41.34 & 23.67 \\
    & & \textit{xlm-roberta-large-ViT-H-14}& 99.09 & 12.51 & 8.49 & 8.63 & 30.25 \\
    & & \textit{ViT-L-14 (laion2b)} &  99.69 & 60.13 & 28.18  & 49.20 & 54.92\\
    & & \textit{ViT-L-14 (openai)} & 96.26 & 70.36 & 44.68 & 36.7 & 48.1 \\
    & & \textit{ViT-B-32 (openai)} & 96.79 & 30.71 & 15.25 & 12.58 & 41.30 \\
    & & \textit{NegCLIP} & 99.35 & 32.26 & 22.22 & 16.39 & 62.63 \\

    \cmidrule{2-8}
    & \multirow{8}{*}{ComCO}    
    &  \textit{ViT-H-14 (DFN)}& 97.45 & 43.49 & 29.20 & 17.91 & 1.13 \\
    & & \textit{ViT-SO400M-SigLIP} & 98.46 & 45.21 & 32.54 & 26.64 & 1.18 \\
    & & \textit{ViT-L-14 (datacomp)}& 92.76 & 40.83 & 17.56 & 9.8 & 1.05 \\
    & & \textit{xlm-roberta-large-ViT-H-14}& 99.84 & 13.18 & 11.02 & 8.26 & 45.38 \\
    & & \textit{ViT-L-14 (laion2b)} & 97.39 & 41.48 & 19.5 & 9.4 & 1.26 \\
    & & \textit{ViT-L-14 (openai)} & 92.81 & 68.46 & 31.85 & 9.8 & 1.24 \\
    & & \textit{ViT-B-32 (openai)} & 95.85 & 42.62 & 22.24 & 9.18 & 0.9 \\
    & & \textit{NegCLIP} & 99.16 & 27.60 & 19.78 & 21.80 & 69.08 \\

    \bottomrule
    \end{tabular}
\end{table*}

\subsection{Text-based Object Retrieval for Long Caption}
\label{app:tor-long}
\begin{table*}[htbp]
    \centering
    \scriptsize
    \setlength{\tabcolsep}{3pt}
    \renewcommand{\arraystretch}{1.1}
    \caption{Text-based Object Retrieval For long template}
    \label{tab:tor_long_total}
    \begin{tabular}{lllcccccc}
    \toprule
    \rowcolor[HTML]{E4E8F2}
    Number of Objects & Dataset & Model & \textbf{Accuracy} & \textbf{First Object} & \textbf{Second Object} & \textbf{Third Object} & \textbf{Fourth Object} & \textbf{Fifth Object} \\ 
    \midrule
    \multirow{16}{*}{n = 2}  & \multirow{8}{*}{SimCO}       
    &  \textit{ViT-H-14 (DFN)}& 96.73 & 62.16 & 37.84 & - & - & -\\
    & & \textit{ViT-SO400M-SigLIP} & 5.88 & 100.0 & 0.00 & - & - & -\\
    & & \textit{ViT-L-14 (datacomp)}& 98.04 & 70.67 & 29.33 & - & - & - \\
    & & \textit{xlm-roberta-large-ViT-H-14}& 98.69 & 76.82 & 23.18 & - & - & - \\
    & & \textit{ViT-L-14 (laion2b)} & 51.63 & 62.03 & 37.97 & - & - & - \\
    & & \textit{ViT-L-14 (openai)} & 96.08 & 39.46 & 60.54 & - & - & - \\
    & & \textit{ViT-B-32 (openai)} & 79.74 & 45.90 & 54.10 & - & - & - \\
    & & \textit{NegCLIP} & 99.35 & 38.82 & 61.18 & - & - & -\\

    \cmidrule{2-9}
    
    & \multirow{8}{*}{ComCO}    
    &  \textit{ViT-H-14 (DFN)}& 92.38 & 71.03 & 28.97 & - & - & -\\
    & & \textit{ViT-SO400M-SigLIP} & 3.42 & 100.0 & 0.00 & - & - & -\\
    & & \textit{ViT-L-14 (datacomp)}& 84.32 & 62.63 & 37.37 & - & - & -\\
    & & \textit{xlm-roberta-large-ViT-H-14}& 72.06 & 63.31 & 36.69 & - & - & -\\
    & & \textit{ViT-L-14 (laion2b)} & 58.73 & 63.01 & 36.99 & - & - & -\\
    & & \textit{ViT-L-14 (openai)} & 84.64 & 61.27 & 38.70 & - & - & -\\
    & & \textit{ViT-B-32 (openai)} & 78.38 & 61.77 & 37.78 & - & - & -\\
    & & \textit{NegCLIP} & 82.67 & 55.63 & 44.37 & - & - & -\\
    
    \midrule 
    \multirow{16}{*}{n = 3} & \multirow{8}{*}{SimCO}   
    &  \textit{ViT-H-14 (DFN)}& 88.6 & 43.02 & 30.43 & 26.56 & -  & -\\
    & & \textit{ViT-SO400M-SigLIP} & 0.74 & 100.0 & 0.00 & 0.00 & -  & -\\
    & & \textit{ViT-L-14 (datacomp)}& 88.48 & 63.02 & 24.38 & 12.60 & -  & -\\
    & & \textit{xlm-roberta-large-ViT-H-14}& 89.83 & 61.66 & 22.10 & 16.23 & -  & -\\
    & & \textit{ViT-L-14 (laion2b)} & 31.86 & 56.54 & 26.15 & 17.31 & -  & -\\
    & & \textit{ViT-L-14 (openai)} & 69.73 & 24.08 & 39.89 & 36.03 & -  & -\\
    & & \textit{ViT-B-32 (openai)} & 38.24 & 25.96 & 39.10 & 34.94 & -  & -\\
    & & \textit{NegCLIP} & 72.30 & 23.39 & 52.71 & 23.90 & -  & -\\

    \cmidrule{2-9}
    & \multirow{8}{*}{ComCO}    
    &  \textit{ViT-H-14 (DFN)}& 76.75 & 50.43 & 22.45 & 27.12 & - & -\\
    & & \textit{ViT-SO400M-SigLIP} & 0.07 & 100.0 & 0.00 & 0.00 & - & -\\
    & & \textit{ViT-L-14 (datacomp)}& 56.14 & 47.80 & 34.17 & 18.03 & - & -\\
    & & \textit{xlm-roberta-large-ViT-H-14}& 36.78 & 48.46 & 28.75 & 22.79 & - & -\\
    & & \textit{ViT-L-14 (laion2b)} & 29.17 & 48.75 & 35.78 & 15.47 & - & -\\
    & & \textit{ViT-L-14 (openai)} & 52.38 & 43.44 & 37.00 & 19.53 & - & -\\
    & & \textit{ViT-B-32 (openai)} & 49.97 & 47.58 & 30.75 & 21.45 & - & -\\
    & & \textit{NegCLIP} & 50.80 & 38.67 & 38.16 & 23.17 & - & -\\
     
    \midrule
    \multirow{16}{*}{n = 4} & \multirow{8}{*}{SimCO}   
    &  \textit{ViT-H-14 (DFN)}& 66.47 & 39.82 & 21.88 & 24.34 & 13.96  & -\\
    & & \textit{ViT-SO400M-SigLIP} & 0.49 & 100.0 & 0.00 & 0.00 & 0.00   & -\\
    & & \textit{ViT-L-14 (datacomp)}& 74.58 & 61.74 & 22.17 & 10.96 & 5.13  & -\\
    & & \textit{xlm-roberta-large-ViT-H-14}& 65.95 & 53.96 & 21.36 & 19.33 & 5.35  & -\\
    & & \textit{ViT-L-14 (laion2b)} & 22.42 & 66.76 & 17.78 & 11.22 & 4.23  & -\\
    & & \textit{ViT-L-14 (openai)} & 58.73 & 16.30 & 32.78 & 26.49 & 24.37  & -\\
    & & \textit{ViT-B-32 (openai)} & 18.43 & 35.64 & 37.77 & 14.18 & 12.41  & -\\
    & & \textit{NegCLIP} & 50.78 & 26.25 & 49.94 & 16.73 & 7.08  & -\\

    \cmidrule{2-9}
    & \multirow{8}{*}{ComCO}    
    &  \textit{ViT-H-14 (DFN)}& 52.87 & 47.87 & 20.54 & 22.72 & 8.87 & -\\
    & & \textit{ViT-SO400M-SigLIP} & 0.01 & 100.0 & 0.00 & 0.00 & 0.00 & -\\
    & & \textit{ViT-L-14 (datacomp)}& 31.36 & 39.21 & 30.74 & 20.94 & 9.11 & -\\
    & & \textit{xlm-roberta-large-ViT-H-14}& 14.99 & 43.03 & 24.29 & 19.72 & 12.96 & -\\
    & & \textit{ViT-L-14 (laion2b)} & 10.19 & 42.66 & 34.16 & 17.09 & 6.09 & -\\
    & & \textit{ViT-L-14 (openai)} & 28.78 & 35.25 & 31.55 & 19.19 & 13.86 & -\\
    & & \textit{ViT-B-32 (openai)} & 21.62 & 43.69 & 24.57 & 16.78 & 14.59 & -\\
    & & \textit{NegCLIP} & 19.41 & 30.36 & 30.38 & 24.39 & 14.86 & -\\
     
    \midrule
    \multirow{16}{*}{n = 5}& \multirow{8}{*}{SimCO}   
    
    &  \textit{ViT-H-14 (DFN)}& 45.44 & 43.46 & 20.45 & 18.34 & 11.87 & 5.88 \\
    & & \textit{ViT-SO400M-SigLIP} & 0.16 & 100.0 & 0.00 & 0.00 & 0.00 & 0.00 \\
    & & \textit{ViT-L-14 (datacomp)}& 51.45 & 59.26 & 22.46 & 8.12 & 8.46 & 1.70 \\
    & & \textit{xlm-roberta-large-ViT-H-14}& 52.92 & 54.87 & 13.81 & 19.30 & 8.16 & 3.86 \\
    & & \textit{ViT-L-14 (laion2b)} & 12.34 & 75.40 & 10.31 & 8.42 & 4.26 & 1.61 \\
    & & \textit{ViT-L-14 (openai)} & 29.39 & 8.98 & 29.39 & 28.44 & 15.97 & 17.20 \\
    & & \textit{ViT-B-32 (openai)} & 6.69 & 32.11 & 38.57 & 12.22 & 8.55 & 8.55 \\
    & & \textit{NegCLIP} & 17.54 & 23.15 & 41.18 & 24.48 & 7.65 & 3.53 \\

    \cmidrule{2-9}
    & \multirow{8}{*}{ComCO}    
    &  \textit{ViT-H-14 (DFN)}& 23.56 & 36.07 & 19.21 & 22.65 & 11.90 & 10.17 \\
    & & \textit{ViT-SO400M-SigLIP} & 0.00 & 100.0 & 0.00 & 0.00 & 0.00 & 0.00 \\
    & & \textit{ViT-L-14 (datacomp)}& 12.49 & 32.55 & 27.84 & 23.76 & 12.73 & 3.11 \\
    & & \textit{xlm-roberta-large-ViT-H-14}& 9.26 & 40.26 & 21.35 & 18.16 & 11.99 & 8.23 \\
    & & \textit{ViT-L-14 (laion2b)} & 4.57 & 38.49 & 31.50 & 17.50 & 8.31 & 4.20 \\
    & & \textit{ViT-L-14 (openai)} & 1.75 & 21.59 & 18.57 & 20.25 & 20.54 & 19.02 \\
    & & \textit{ViT-B-32 (openai)} & 1.86 & 32.72 & 15.62 & 14.71 & 18.36 & 16.26 \\
    & & \textit{NegCLIP} & 1.41 & 24.30 & 23.17 & 22.14 & 17.64 & 12.75 \\

    \bottomrule
    \end{tabular}
    \end{table*}

In this section, we aimed to examine the performance of various models in the IOR experiment when presented with longer caption formats. This approach mirrors our previous investigation, allowing us to draw comparisons between standard and extended caption scenarios.

We utilized the same extended caption template as in the previous section.
The results of this experiment are presented in Table \ref{tab:tor_long_total}. Notably, the observed pattern closely aligns with that of the standard IOR experiment, suggesting a consistency in model behavior across different caption lengths.


\clearpage

\subsection{LAION Dataset Analysis}
\label{subsec:laion_analysis}
\begin{figure*}[h]
    \centering
    \includegraphics[width=\linewidth]{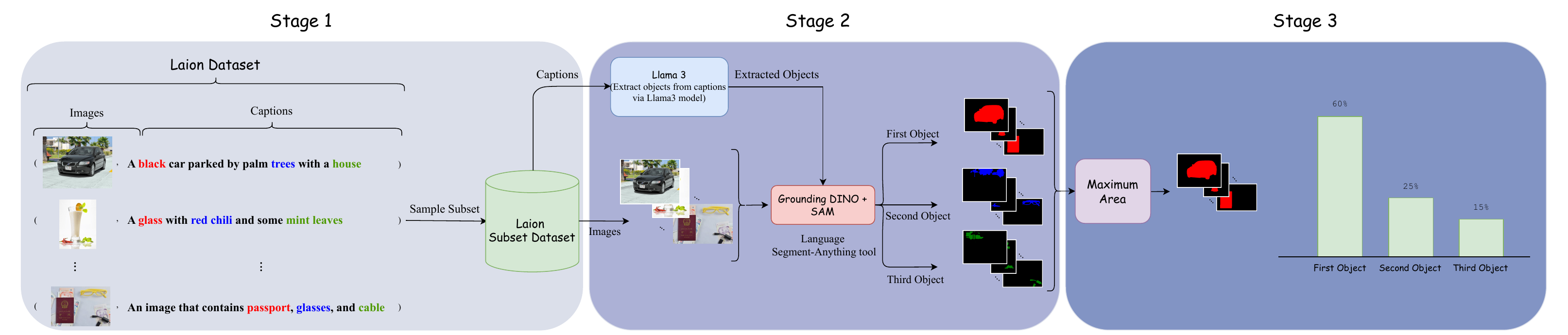}
    \caption{Process flow for LAION dataset analysis}
    \label{fig:laion-objects}
\end{figure*}

To investigate the potential bias in CLIP's training data, as discussed in Section 4.3, Claim 2, we conducted an analysis of the LAION dataset. This process, illustrated in Figure \ref{fig:laion-objects}, consisted of three main stages:

\subsubsection{Stage 1: Dataset Sampling}

Due to the vast size of the LAION dataset (over 2 billion image-text pairs), we randomly selected a subset of 200,000 samples for our analysis. This subset maintained the diversity of the original dataset while making the analysis computationally feasible.

\subsubsection{Stage 2: Object Extraction}

For each image-caption pair in our subset:

\begin{enumerate}
    \item We used the Llama 3 model to extract object mentions from the captions. This step allowed us to identify the objects described in each text without relying on manual annotation.
    
    \item We applied the Grounding DINO + SAM (Segment Anything Model) tool to generate object masks for the corresponding images. This process enabled us to identify and segment individual objects within each image.
\end{enumerate}

\subsubsection{Stage 3: Analysis}

With the extracted data, we performed the following analysis:

\begin{enumerate}
    \item \textbf{Object Order:} We recorded the order in which objects were mentioned in each caption.
    
    \item \textbf{Object Size:} Using the generated masks, we calculated the area of each object in the corresponding image.
    
    \item \textbf{Correlation:} We examined the relationship between an object's position in the caption and its size in the image.
\end{enumerate}



AS shown in Figure \ref{fig:coco_analysis_total} This distribution strongly suggests a bias in the LAION dataset where larger objects tend to be mentioned earlier in image captions. This finding supports our hypothesis about the origin of CLIP's text encoder bias, as discussed in Section 4.3 of the main paper.

\subsection{COCO Dataset Analysis}
\label{app:coco-anlysis}
In this section, we repeated the experiment conducted in Section 4.3 for different scenarios involving 2 to 5 objects. We divided the captions in the COCO dataset into four subsets: those mentioning 2 objects, 3 objects, 4 objects, and 5 objects. We then analyzed each subset to determine in what percentage of cases the largest object appeared in which position.

\begin{figure*} [h!]
    \centering
    \includegraphics[width=\linewidth]{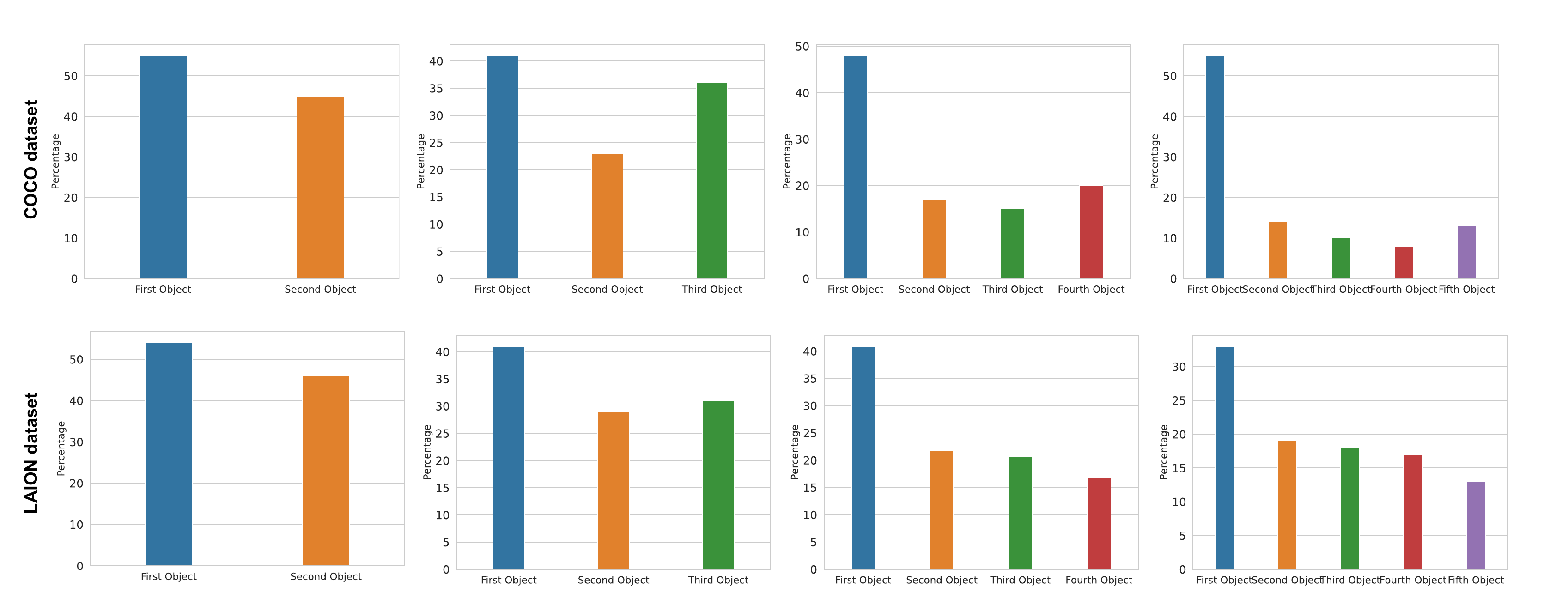}
    \caption{Distribution of larger object positions in captions for objects in COCO and LAION dataset}
    \label{fig:coco_analysis_total}
\end{figure*}

The results of this evaluation are presented in Figure \ref{fig:coco_analysis_total}. As can be observed, this trend is repeated across all scenarios: in most cases, the larger object appears earlier in the caption.

\clearpage

\subsection{Object Categories from DomainNet}
\label{app:categorized_domainnet}
The DomainNet dataset objects were categorized into three groups based on their relative sizes: small, medium, and large. These categories were used to investigate potential bias in CLIP's text embeddings, as discussed in Section 4.3, Claim 1. The full list of objects used in each category is presented below:

\subsubsection{Small Objects}
\begin{center}
\scriptsize 
\begin{tabular}{@{} p{2cm} p{2cm} p{2cm} p{2cm} @{}}
ant & anvil & apple & arm \\
asparagus & axe & banana & bandage \\
basket & bat & bee & belt \\
binoculars & bird & blackberry & blueberry \\
book & boomerang & bottlecap & bowtie \\
bracelet & brain & bread & broccoli \\
broom & bucket & butterfly & cactus \\
cake & calculator & calendar & camera \\
candle & carrot & cat & clarinet \\
clock & compass & cookie & crab \\
backpack & crown & cup & dog \\
donut & drill & duck & dumbbell \\
ear & envelope & eraser & eye \\
eyeglasses & feather & finger & fork \\
frog & hammer & hat & headphones \\
hedgehog & helmet & hourglass & jacket \\
keyboard & key & knife & lantern \\
laptop & leaf & lipstick & lobster \\
lollipop & mailbox & marker & megaphone \\
microphone & microwave & mosquito & mouse \\
mug & mushroom & necklace & onion \\
owl & paintbrush & parrot & peanut \\
pear & peas & pencil & pillow \\
pineapple & pizza & pliers & popsicle \\
postcard & potato & purse & rabbit \\
raccoon & radio & rake & rhinoceros \\
rifle & sandwich & saw & saxophone \\
scissors & scorpion & shoe & shovel \\
skateboard & skull & snail & snake \\
snorkel & spider & spoon & squirrel \\
stethoscope & strawberry & swan & sword \\
syringe & teapot & telephone & toaster \\
toothbrush & trombone & trumpet & umbrella \\
violin & watermelon & wheel & \\
\end{tabular}
\end{center}

\subsubsection{Medium Objects}
\begin{center}
\scriptsize
\begin{tabular}{@{} p{2cm} p{2cm} p{2cm} p{2cm} @{}}
angel & bathtub & bear & bed \\
bench & bicycle & camel & cannon \\
canoe & cello & chair & chandelier \\
computer & cooler & couch & cow \\
crocodile & dishwasher & dolphin & door \\
dresser & drums & flamingo & guitar \\
horse & kangaroo & ladder & mermaid \\
motorbike & panda & penguin & piano \\
pig & sheep & stereo & stove \\
table & television & tiger & zebra \\
\end{tabular}
\end{center}

\subsubsection{Large Objects}
\begin{center}
\scriptsize
\begin{tabular}{@{} p{2cm} p{2cm} p{2cm} p{2cm} @{}}
aircraft carrier & airplane & ambulance & barn \\
bridge & bulldozer & bus & car \\
castle & church & cloud & cruise ship \\
dragon & elephant & firetruck & flying saucer \\
giraffe & helicopter & hospital & hot air balloon \\
house & moon & mountain & palm tree \\
parachute & pickup truck & police car & sailboat \\
school bus & skyscraper & speedboat & submarine \\
sun & tent & The Eiffel Tower & Wall of China \\
tractor & train & tree & truck \\
van & whale & windmill & \\
\end{tabular}
\end{center}

\subsection{Text to image generation}
\label{sec:appendix_text_to_image}

The biases observed in CLIP's encoders have significant implications beyond image-text matching, particularly for text-to-image generation models that incorporate CLIP components. To investigate this impact, we focused on Stable Diffusion, a popular text-to-image generation model that utilizes CLIP's text encoder in its pipeline.
Stable Diffusion employs CLIP's text encoder to process input prompts, creating text embeddings that guide the image generation process. Given our identification of biases in CLIP's text encoder, especially the preference for objects mentioned earlier in text descriptions, we hypothesized that these biases would manifest in the generated images.
To test this hypothesis, we designed an experiment using prompts containing multiple objects from the COCO dataset. Our goal was to observe whether the order of objects in the text prompt influences their prominence or likelihood of appearance in the generated images.

Our experimental methodology consisted of three main steps. First, we created 1,000 multi-object prompts, each containing four distinct objects from the COCO dataset. Second, we used these prompts to generate images using three versions of Stable Diffusion: v1.4 \cite{Rombach_2022_CVPR}, v2, and SD-XL \cite{podell2023sdxl}. Finally, to evaluate the presence of objects in the generated images, we employed YOLO v8 \cite{reis2023real}, a state-of-the-art object detection model. We configured YOLO v8 with a detection threshold of 0.25 and used it to validate which objects from the original prompt were present in the generated image.

This approach allowed us to quantitatively assess how CLIP's text encoder biases propagate through the Stable Diffusion pipeline and manifest in the generated images. By comparing the frequency of object detection with their position in the input prompt, we could directly observe the impact of the text-side bias on the image generation process.

\begin{table}[ht]
\centering
\scriptsize
\setlength{\tabcolsep}{3pt}
\renewcommand{\arraystretch}{1.2}
\caption{Object presence in Stable Diffusion-generated images }
\label{tab:image_gen}
\begin{tabular}{lcccc}
\toprule
\rowcolor[HTML]{EFEFEF}
Model & \textbf{First Obj} & \textbf{Second Obj} & \textbf{Third Obj} & \textbf{Fourth Obj} \\ 
\midrule 
\textit{SD v1.4} & 57.7 & 44.7 & 38.1 & 35.4 \\
\textit{SD V2} & 62.5 & 49.7 & 47.5 & 42.2 \\
\textit{SD-XL}  & 79.2 & 69.3 & 59.4 & 64.0 \\
\bottomrule
\end{tabular}
\end{table}

Our findings, presented in Table \ref{tab:image_gen}, demonstrate a clear correlation between an object's position in the text prompt and its likelihood of appearing in the generated image. This correlation aligns with our earlier observations of CLIP's text encoder bias, suggesting that these biases significantly influence the output of text-to-image generation models.

\subsection{Preliminary Method for Bias Mitigation}
\label{sec:pre-method}
In our analysis, we observed a critical limitation in the text encoder of CLIP: it disproportionately prioritizes objects mentioned earlier in captions. This bias results in embeddings that heavily represent the first object while progressively diminishing the contribution of subsequent objects. To mitigate this, we explored a novel strategy to reduce positional dependence in object representations.
\subsubsection{Proposed Solution}

We propose splitting a given caption into multiple sub-captions, each focusing on a single object. By generating embeddings for each sub-caption and aggregating these embeddings, we aim to achieve a balanced representation that minimizes positional bias.

To evaluate this approach, we utilized the ComCO dataset, where objects in captions are separated by the conjunction \textit{`and'}. This structure allowed straightforward decomposition of captions into sub-captions corresponding to individual objects. We conducted the image-text matching experiment (described in Section~\ref{sec:impact}) under two conditions: (1) using original captions as-is and (2) using the aggregated embeddings from split captions. Results from this comparison are presented in Table~\ref{table:imagetext_matching_split}.

\subsubsection{Results and Observations}

As shown in Table~\ref{table:imagetext_matching_split}, the aggregated approach led to a substantial improvement in image-text matching accuracy. This outcome suggests that reducing the influence of positional bias can enhance the text encoder's performance in multi-object scenarios. Our findings further underscore the potential of designing methods that neutralize word order effects, thereby enabling more robust and unbiased embeddings.

\begin{table}[ht]
\centering
\scriptsize
\setlength{\tabcolsep}{7pt} 
\renewcommand{\arraystretch}{1.2} 

\caption{Image-Text Matching Accuracy for ComCO Dataset with Original and Split Caption Aggregation Approaches. The first scenario represents results using original captions, while the second scenario reflects the aggregated embeddings of split captions.}
\label{table:imagetext_matching_split}
\begin{tabular}{l c c}
\toprule
\rowcolor[HTML]{EFEFEF}
\textbf{Model} & \textbf{Original Captions} (\%) & \textbf{Split Caption Aggregation} (\%) \\ 
\midrule

\textit{CLIP Datacomp} \cite{gadre2024datacomp} & 67.50 & \textbf{98.39} \\
\textit{CLIP Roberta} & 64.75 & \textbf{97.35} \\
\textit{SIGLIP} \cite{zhai2023sigmoid} & 72.36 & \textbf{99.05} \\
\textit{CLIP openAI} & 52.23 & \textbf{88.56} \\
\textit{NegCLIP} & 46.94 & \textbf{96.82} \\

\bottomrule
\end{tabular}
\end{table}

\subsubsection{Limitations and Future Directions}

We acknowledge that this solution, while effective for the ComCO dataset, is a heuristic and dataset-specific approach. Its generalizability remains limited. Nonetheless, this experiment demonstrates our commitment to exploring practical solutions and provides a foundation for future advancements.

Future work will focus on developing scalable methods to address positional bias. Possible directions include leveraging large language models (LLMs) to automate caption decomposition into sub-captions and modifying the positional embeddings in the text encoder to ensure equal representation of all objects. These efforts aim to provide a more comprehensive and generalizable solution, paving the way for improved robustness in vision-language models.

\end{document}